\documentclass{article}

\usepackage{microtype}
\usepackage{graphicx}
\usepackage{subcaption}
\usepackage{booktabs} 

\usepackage{hyperref}
\usepackage{useful_math}
\usepackage{xcolor}
\usepackage{mathrsfs}
\usepackage{multirow}
\usepackage{stfloats}
\usepackage{algorithm}
\usepackage{listings}
\usepackage{float}
\usepackage{natbib}

\usepackage[preprint]{icml2026}

\makeatletter
\renewcommand{\ICML@preprint}{\vspace{-2.5em}} 
\makeatother

\usepackage{amsmath}
\usepackage{amssymb}
\usepackage{mathtools}
\usepackage{amsthm}
\usepackage{xspace}  
\makeatletter
\DeclareRobustCommand\onedot{\futurelet\@let@token\@onedot}
\def\@onedot{\ifx\@let@token.\else.\null\fi\xspace}

\def\eg{\emph{e.g}\onedot} 
\def\ie{\emph{i.e}\onedot}

\makeatother

\usepackage[capitalize,noabbrev]{cleveref}

\theoremstyle{plain}

\theoremstyle{definition}

\theoremstyle{remark}

\newcommand{\eqcomma}[0]{\;,}
\newcommand{\eqdot}[0]{\;.}

\newcommand{\tinycite}[1]{{\tiny{\cite{#1}}}}
\newcommand{\tinycitepalias}[1]{{\tiny{\citepalias{#1}}}}

\usepackage{etoolbox}
\makeatletter
\AfterEndEnvironment{algorithm}{\let\@algcomment\relax}
\AtEndEnvironment{algorithm}{\kern2pt\hrule\relax\vskip3pt\@algcomment}
\let\@algcomment\relax
\newcommand\algcomment[1]{\def\@algcomment{\footnotesize#1}}
\renewcommand\fs@ruled{\def\@fs@cfont{\bfseries}\let\@fs@capt\floatc@ruled
  \def\@fs@pre{\vspace{0.4em}\hrule height.8pt depth0pt \kern2pt}%
  \def\@fs@post{\vspace{-1.0em}}
  \def\@fs@mid{\kern2pt\hrule\kern2pt}%
  \let\@fs@iftopcapt\iftrue}
\makeatother

\usepackage[textsize=tiny]{todonotes}

\icmltitlerunning{Image Generation with a Sphere Encoder}

\begin{document}

\twocolumn[
  \icmltitle{
    Image Generation with a Sphere Encoder
  }



  \icmlsetsymbol{equal}{$\dag$}
  \icmlsetsymbol{intern}{\textsection}

  \begin{icmlauthorlist}
    \icmlauthor{Kaiyu Yue}{equal,comp,sch}
    \icmlauthor{Menglin Jia}{equal,comp}
    \icmlauthor{Ji Hou}{comp}
    \icmlauthor{Tom Goldstein}{sch}
  \end{icmlauthorlist}

  \icmlaffiliation{comp}{Meta\;}
  \icmlaffiliation{sch}{University of Maryland}


  \icmlkeywords{Sphere Encoders, Few-step Generation, One-step Generation, Autoencoders, Generative Models}
  \vskip 0.3in
]



\printAffiliationsAndNotice{
  $^\dag$Equal contribution.
}
\begin{abstract}
  We introduce the Sphere Encoder, an efficient generative framework capable of producing images in a single forward pass and competing with many-step diffusion models using fewer than five steps.
  Our approach works by learning an encoder that maps natural images uniformly onto a spherical latent space, and a decoder that maps random latent vectors back to the image space.
  Trained solely through image reconstruction losses, the model generates an image by simply decoding a random point on the sphere.
  Our architecture naturally supports conditional generation, and looping the encoder/decoder a few times can further enhance image quality.  Across  several datasets, the sphere encoder approach yields performance competitive with state of the art diffusions, but with a small fraction of the inference cost.
  Project page is available at \href{https://sphere-encoder.github.io}{\texttt{sphere-encoder.github.io}}.
\end{abstract}

\begin{figure}[t!]
  \centering
  \centerline{\includegraphics[width=1.0\linewidth]{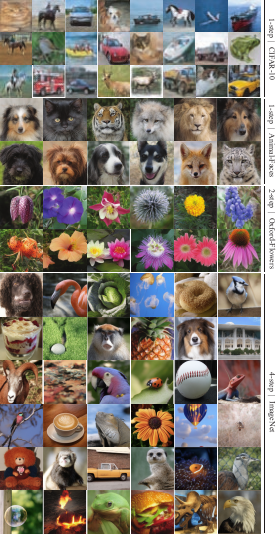}}
  \caption{
    \textbf{Selected images generated by the Sphere Encoder} in one-step for CIFAR-10 ($32\times32$) and Animal-Faces, two-steps for Oxford-Flowers, and four-steps for ImageNet ($256\times256$).
  }
  \label{fig:teaser}
  \vspace{-2.em}
\end{figure}

\section{Introduction}
\label{sec:intro}

Most generative image models rely on either diffusion \cite{ddpm,fm} or autoregressive next-token prediction \cite{tian2024visual}. With either paradigm, image generation is extremely slow and costly, requiring many forward passes to produce a single image.

\begin{figure}[t]
  \centering
  \centerline{\includegraphics[width=0.8\columnwidth]{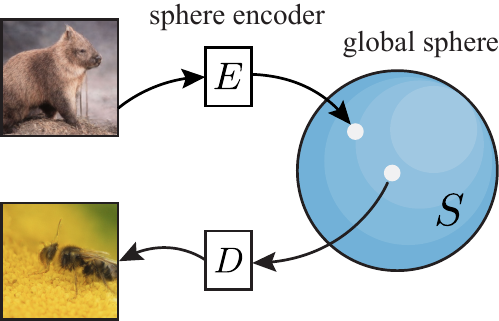}}
  \caption{
    {\bf{A sphere encoder}} $E$ maps the natural image distribution uniformly onto a global sphere $S$.
    The decoder $D$ then generates a realistic image by decoding a random point on the sphere.
  }
  \label{fig:intro}
  \vspace{-1.0em}
\end{figure}

We propose an alternative paradigm that is capable of generating sharp images with as little as one forward pass. Our approach, which we call a {\em sphere encoder}, works by training two complementary models: an encoder model that maps the distribution of natural images uniformly onto the sphere, and a decoder that maps points on the sphere back to natural images (\cref{fig:intro}).
The term aligns with the autoencoder convention, reflecting its encoder-decoder architecture.
At test time, an image is generated quickly by sampling a random point on the sphere and passing it through the decoder.

Although the sphere encoder does not employ diffusion processes explicitly, it supports several key capabilities commonly associated with its diffusion-based cousins \cite{add,ldm,stable-diffusion-3}.  These include conditional generation using AdaLN \cite{adaLN,dit}, classifier-free guidance (CFG) \cite{cfg}, and few-step iteration to enhance sample quality \cite{gan,glow,consistency-models}.
Experiments demonstrate that our approach achieves competitive one-step generation, and state-of-the-art performance in few-step regimes (\eg, fewer than $5$ steps) on a range of datasets.

\subsection*{Motivation and Relation to Autoencoders}

Autoencoders \cite{autoencoders-lecun,autoencoders-bourlard,autoencoders-hinton} have been widely used in representation learning and generative modeling.
A lower-dimensional latent bottleneck between the encoder and decoder forces the model to learn an undercomplete representation of the input \cite{dl-book}.

\begin{figure}[t]
  \centering
  \centerline{\includegraphics[width=0.9\linewidth]{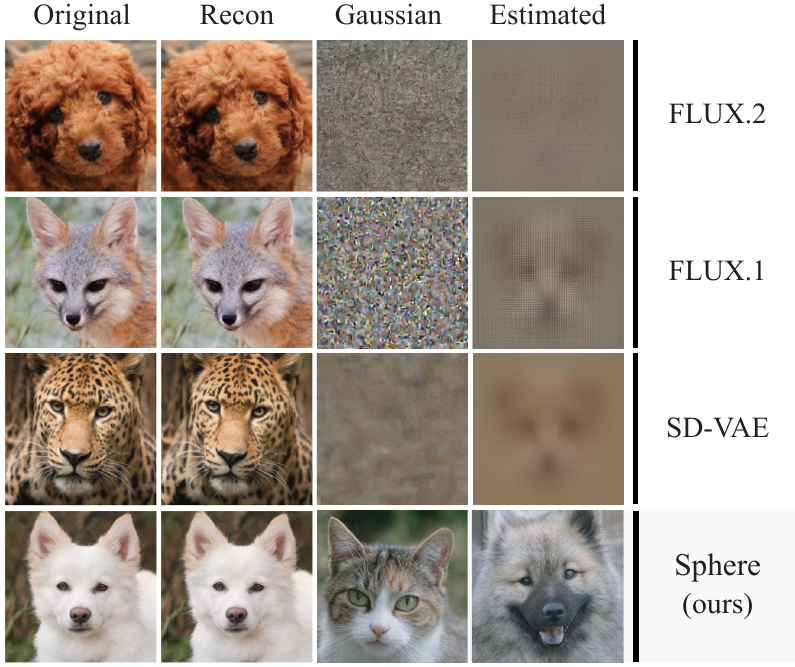}}
  \caption{
    \textbf{Posterior hole problem in VAEs}.
    Columns:
    (1) Input images;
    (2) Autoencoder reconstructions;
    (3) Samples from standard Gaussian prior;
    and (4) Samples from estimated Gaussian posterior on Animal-Faces training set.
    Unlike modern FLUX.1/2 \cite{flux} and SD-VAE \cite{sdxl}, our sphere encoder produces realistic images by decoding random points sampled from the sphere.
  }
  \label{fig:rsp}
  \vspace{-1.0em}
\end{figure}

To regularize the latent space, variational autoencoders (VAEs) \cite{vae,vae-introduction,wae,sae,sae-kl} minimize the divergence between the latent distribution and a (typically) Gaussian prior.
Unfortunately, in the standard VAE formulation, the divergence loss and image reconstruction loss are at odds with one another; zero divergence loss cannot be achieved simultaneously with perfect image reconstruction. As a result, the learned posterior fails to strongly match the prior -- an issue known as the posterior hole problem \cite{adversarial-ae,taming-vae,vampprior,diagnosing-ae,regularized-ae,contrastive-ae}.
Direct samples from the Gaussian prior fail to yield valid images.
Realistic images are currently possible only by decoding samples from the posterior (\ie, adding noise to latents derived from real images), as illustrated in \cref{fig:rsp}.
Our approach does not suffer from this problem.

Like a classical VAE, our approach relies on an autoencoder.  Unlike the VAE, which tries to force the latent vectors into a Gaussian distribution, we instead force latents to be uniformly distributed on a sphere.
Due to the bounded and rotationally symmetric nature of the sphere, this can be achieved simply by forcing embeddings of natural images away from one another, causing them to spread throughout the sphere.
Moreover, this objective is not in contradiction with the image reconstruction objective; we can achieve both uniformity and accurate reconstruction simultaneously.

Many contemporary state-of-the-art diffusion models are actually latent diffusion models \cite{ldm,dit,rectified-flow,sit,stable-diffusion-3,sdxl,wan} -- hybrid models built on top of VAEs. The VAE partially Gaussianizes the image distribution, but not well enough to be sampled.  A diffusion pipeline picks up the slack in the VAE, going the last mile of producing a valid latent sample for the decoder.
Concurrent works have shown that more powerful representation encoders \cite{repa,tong2026scaling}, and even spherical manifold encoders \cite{rae}, result in faster training of the diffusion layer. In our work, we show that a spherical latent space\footnote{
  In contrast to prior vMF-based approaches, we create our spherical space using simple vector RMS normalization.
} can be learned so precisely that the expensive diffusion step is irrelevant.

\section{Method}
\label{sec:method}

\subsection{Spherical Latent Space}
\label{sec:spherical_latent_space}

We employ an encoder $E$ based on a Transformer \cite{vit,transformer} to map an input image $\x \in \real^{H\times W\times3}$ into a latent representation $\z \in \real^{h\times w\times d}$.
The latent resolution is determined by the patch size $P$, such that $h = H / P$ and $w = W / P$, with $d$ denoting the channel depth.

To construct a global spherical latent space, we define a \emph{spherifying} function, denoted as $f$.
This function flattens $\z$ into a vector of dimension $L = h \times w \times d$ and then projects it onto a sphere with radius $\sqrt{L}$ via RMS normalization:
\begin{align}
  \label{eq:naive_spherify}
  \bv = f(\z) = f(E(\x))\eqdot
\end{align}
Subsequently, a decoder $D$ reconstructs the image from $\bv$:
\begin{align}
  \label{eq:reconstruction}
  \hat{\x} = D(\bv)\eqcomma
\end{align}
where $\hat{\x}$ denotes the reconstructed image.
If the encoder maps images uniformly onto a sphere, then we can generate images by decoding random points on the sphere:
\begin{align}
  \label{eq:generation}
  \hat{\x} = D(f(\be))\eqcomma
\end{align}
where $\be \sim \mathcal{N}(0, \mathbf{I}) \in \real^L$ is random anisotropic Gaussian and $f(\be)$ is uniformly distributed on the sphere.
For simplicity, we use $\hat{\x}$ to denote the decoder output in both reconstruction and generation scenarios.

\begin{figure*}[t]
  \centering
  \centerline{\includegraphics[width=0.85\linewidth]{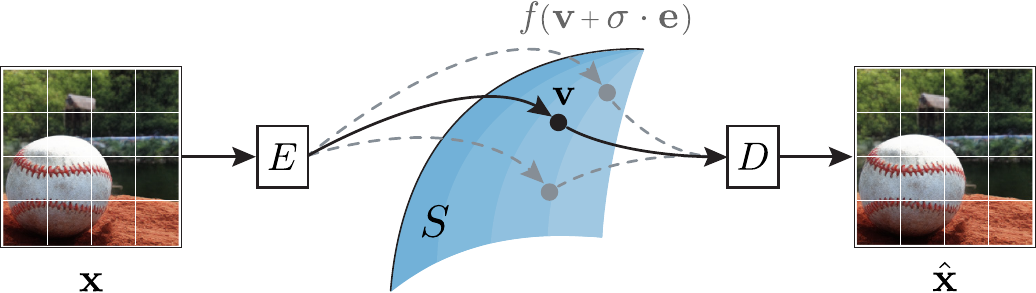}}
  \vspace{0.5em}
  \caption{
    \textbf{Spherifying latent with noise}.
    Encoder $E$ maps image $\x$ to a latent, which $f$ projects to $\bv$ on sphere $S$.
    During training, random Gaussian noise $\sigma \cdot \be$ is added to $\bv$, where $\sigma$ is jittered magnitude.
    Decoder $D$ reconstructs the image $\hat{\x}$ from the re-projected noisy latent $f(\bv + \sigma \cdot \be)$.
  }
  \label{fig:model}
  \vspace{-1.0em}
\end{figure*}

\subsection{Spherifying with Noise}

Our training process uses embedding vectors of natural images, and also noisy versions of those embedding vectors.
The purpose of training with noisy vectors is two-fold.
First, noisy clouds of vectors densely cover the latent space, enabling us to train the decoder on the continuous global latent sphere, rather than only on the finite set of embedding vectors.
Second, by using a loss that promotes accurate decoding of noisy latent vectors, we force the noisy clouds produced by each training image to spread apart and cover the entire latent sphere.

From a Normal distribution, we randomly sample a noise vector $\be \sim \mathcal{N}(0, \mathbf{I}) \in \real^L$ to perturb the direction of $\bv$:
\begin{align}
  \label{eq:noisy_spherify}
  \bv_\text{NOISY} = f( \bv + \sigma \cdot \be )\eqcomma
\end{align}
where the scalar $\sigma$ controls the noise magnitude.
Note that $f$ is applied again here to project the perturbed vector back onto the spherical surface.

\textbf{Jittering Sigma}.
To cover diverse directions on the sphere, we jitter $\sigma$ during the training.
By sampling a scalar $r$ uniformly from $[0, 1]$, we compute $\sigma$ as:
\begin{align}
  \label{eq:sigma_jitter}
  \sigma = r \cdot \sigma_{\text{max}}\eqcomma
\end{align}
where the $\sigma_{\text{max}}$ is the maximum noise limit.
The case of $r=0$ reduces to the naive spherifying in \cref{eq:naive_spherify}.
Later we determine the optimal value for $\sigma_{\text{max}}$ with experiments.
This core design is illustrated in \cref{fig:model}.

\subsection{Training Objective}
\label{sec:losses}

Consider two perturbed latent vectors, $\bv_\text{NOISY}$ and $\bv_\text{noisy}$, with large and small noise.
$\bv_\text{NOISY}$ is defined as in \cref{eq:noisy_spherify} with $\sigma \in [0, \sigma_\text{max}]$.
The other perturbed $\bv_\text{noisy}$ has less jitter:
\begin{align}
  \bv_\text{noisy} = f( \bv + \sigma_\text{sub} \cdot \be )\eqcomma
\end{align}
where $\sigma_\text{sub} = s \cdot \sigma$, and $s$ is uniformly sampled from $[0, 0.5]$.
Note that $\bv_\text{noisy}$ shares the same noise direction $\be$ as $\bv_\text{NOISY}$.

\textbf{Pixel Reconstruction Loss}.
This loss ensures that the decoder is an approximate inverse of the encoder, and that the decoder creates valid images.
We have the standard pixel-level reconstruction loss, which combines of smoothed L1 loss \cite{fast-rcnn} and perceptual loss \cite{perceptual-loss}.
This loss encourages the decoder to reconstruct the input image $\x$ from its noisy latent representation $\bv_\text{noisy}$:
\begin{align}
  \cL_{\text{pix-recon}} = \cL_\text{L1 + perceptual} \left( D(\bv_\text{noisy}), \x \right)\eqdot
\end{align}

\textbf{Pixel Consistency Loss}.
This consistency loss ensures that the latent space is smooth and well structured by promoting that nearby latent vectors produce similar images:
\begin{align}
  \cL_{\text{pix-con}} =
  \cL_\text{L1 + perceptual}(D(\bv_\text{NOISY}), \text{sg}(D(\bv_\text{noisy})))\eqcomma
\end{align}
which also uses the combination of smooth L1 loss and perceptual loss, and $\text{sg}(\cdot)$ denotes stop-gradient operation.

\textbf{Latent Consistency Loss}.
It is well known that image similarity is better measured in latent space than in pixel space \cite{unreasonable-perceptual-loss,clip}.
This is the reason why our pixel similarities use a perceptual loss, which relies on features produced by a static VGG model.
To achieve a stronger consistency loss, we also measure image similarity using the latent space of our own encoder.

We want a natural image $\x$ and its noisy decoded representation $D(\bv_\text{NOISY})$ to be semantically similar.
The semantic similarity is measured by applying the encoder to both, and computing the cosine similarity between their latent representations. This yields the following loss:
\begin{align}
  \cL_{\text{lat-con}} =
  \cL_\text{cosine similarity}( \bv, E(D(\bv_\text{NOISY})))\eqdot
\end{align}
This loss serves an additional important purpose: It improves the iterative generation process we discuss later by encouraging the encoder to map distorted images, $D(\bv_\text{NOISY})$, that may be off the image manifold to ``cleaned up'' latent vectors that reflect on-manifold images.

\textbf{Overall Loss.}
The overall training loss is a weighted sum of the three components:
\begin{align}
  \cL =
  \cL_{\text{pix-recon}}
  + \cL_{\text{pix-con}}
  + \cL_{\text{lat-con}}\eqdot
\end{align}
More details about loss weights and training hyperparameters are provided in Appendix \ref{appsec:hyperparams}.

\subsection{Model Architecture}

Our architecture employs the standard ViT \cite{vit} for both encoder and decoder.
We insert 4-layer MLP-Mixers \cite{mlp-mixer} in the end of the encoder (before spherification) and the beginning of the decoder.
This aims to improve cross-token mixing and globalization of features without the expense of linear layers on the full flattened vector.
A final RMSNorm layer \cite{rmsnorm} with learned affine parameters is added to each MLP-Mixer to bound the latent magnitude ($\leq \sqrt{L}$).
This regularization proves critical for stabilizing training, especially when there is a dramatic divergence between the decoder outputs of $\bv_\text{noisy}$ and $\bv_\text{NOISY}$.
We use both RoPE \cite{roformer} positional embedding and sinusoidal absolute positional encoding.
We found that removing the sinusoidal positional embedding hurts generation quality.

For class-conditional generation, we implement AdaLN-Zero \cite{adaLN,dit} in both the encoder and decoder using separate class embeddings.
A learned {\texttt{null}} embedding is trained for classifier-free guidance (CFG) \cite{cfg} with a probability of $0.1$.
For image reconstruction tasks, we found using either random class embeddings or the {\texttt{null}} embedding is effective in the conditional setting.
We default to the {\texttt{null}} embedding for simplicity.
In addition, CFG can be applied in either the latent space (after the encoder), or the pixel space (after the decoder), or both.

\cref{alg:code} summarizes the forward pass for one-step or few-step generation at inference time.
We count $D(f(\be))$ as one-step generation, and few-step generation as iteratively encoding and decoding $T-1$ times \footnote{
  While a `step' represents a single model iteration regardless of CFG, NFE (number of function evaluation) counts the dual forward passes required by CFG.
}.
We fix $r = 1.0$ in \cref{eq:sigma_jitter} to use the maximum noise magnitude across all steps.
For reconstruction task, no noise is added ($r = 0.0$).

\begin{algorithm}[t]
  \caption{pseudocode of generation forward pass.}
  \label{alg:code}
  \definecolor{codeblue}{rgb}{0.25,0.45,0.70}
  \lstset{
    backgroundcolor=\color{white},
    basicstyle=\fontsize{7.2pt}{7.2pt}\ttfamily\selectfont,
    columns=fullflexible,
    breaklines=true,
    captionpos=b,
    commentstyle=\fontsize{7.2pt}{7.2pt}\color{codeblue},
    keywordstyle=\fontsize{7.2pt}{7.2pt},
  }
  \begin{lstlisting}[language=python]
e = Normal(0, 1).sample([L])  # random noise vector

# one-step generation
v = spherify(e, sampling=False)  # Eq.(1) w/o noise
x = D(v, y)  # y is class embedding

if do_dec_cfg:  # in pixel space
    x_uncond = D(v, y_null)  # null class embedding
    x = x_uncond + cfg * (x - x_uncond)

# few-step refinement
for _ in range(T - 1):  # T steps in total
    z = E(x, y)

    if do_enc_cfg:  # in latent space
        z_uncond = E(x, y_null)
        z = z_uncond + cfg * (z - z_uncond)
    
    v = spherify(z, sampling=True)  # Eq.(4) w/ noise
    x = D(v, y)

    if do_dec_cfg:
        x_uncond = D(v, y_null)
        x = x_uncond + cfg * (x - x_uncond)
        
return x
\end{lstlisting}
\end{algorithm}

\section{Quantitative Experiments}
We adopt generation FID (gFID) \cite{fid} and Inception Score (IS) \cite{inception-score} to assess generation quality, while reconstruction FID (rFID) measures reconstruction quality.
All metrics are computed using $50$K randomly sampled training images.
For class-conditional generation, we have a balanced distribution with an equal number of random samples per class.

We perform experiments on CIFAR-10 \cite{cifar-10} with small image size $32\times32$, as well as ImageNet \cite{imagenet}, Animal-Faces \cite{starganv2}, and Oxford-Flowers \cite{oxford-flowers} with large image size $256\times256$.
Center cropping and horizontal flipping with $0.5$ probability are the only data augmentation.

\begin{table}[t]
  \centering
  \scriptsize
  \scshape
  \caption{
    \textbf{Few-step generation results} on CIFAR-10.
  }
  \label{tab:cifar}
  \setlength{\tabcolsep}{4.8pt} 
  \begin{tabular}{@{}lccrr@{}} 
    \toprule
    method
     & steps & rFID $\downarrow$ & gFID $\downarrow$ & IS $\uparrow$ \\
    \midrule
    \multicolumn{5}{c}{conditional generation w/o cfg}               \\
    \cmidrule{1-5}
    sphere-l
     & 1     & 0.59              & 18.68             & 9.1           \\
     & 2     & -                 & 8.28              & 9.9           \\
     & 4     & -                 & 2.72              & 10.5          \\
     & 6     & -                 & 1.65              & 10.7          \\
    \cmidrule{1-5}
    stylegan{\tiny{2}} \tinycite{style-gan-2}
     & 1     & -                 & 6.96              & 9.53          \\
    stylegan{\tiny{2 + ADA}} \tinycite{style-gan-2}
     & 1     & -                 & 3.49              & 10.2          \\
    \midrule
    \multicolumn{5}{c}{unconditional generation}                     \\
    \cmidrule{1-5}
    sphere-l
     & 1     & 0.53              & 35.67             & 6.7           \\
     & 2     & -                 & 14.13             & 8.4           \\
     & 4     & -                 & 4.31              & 9.8           \\
     & 6     & -                 & 2.34              & 10.2          \\
    \cmidrule{1-5}
    ddim \tinycite{ddim}
     & 1K    & -                 & 3.17              & -             \\
    ddpm \tinycite{ddpm}
     & 1K    & -                 & 3.17              & 9.4           \\
    improved-ddpm \tinycite{improved-ddpm}
     & 4K    & -                 & 2.90              & -             \\
    \bottomrule
  \end{tabular}
  \vspace{-2.05em}
\end{table}

\subsection{Small Image Size}
We first comprehensively test our method with small image size ($32\times32$) on CIFAR-10 \cite{cifar-10}.
We build encoder and decoder with ViT Large, which consists of $24$ layers and yields a latent dimension $L = 16\times16\times8$.
The model, indicated as Sphere-L, is trained for $5000$ epochs for conditional generation, $10000$ epochs for unconditional generation, following the other setup in Appendix \ref{appsec:hyperparams}.

\begin{figure*}[t]
  \centering
  \centerline{\includegraphics[width=1.0\linewidth]{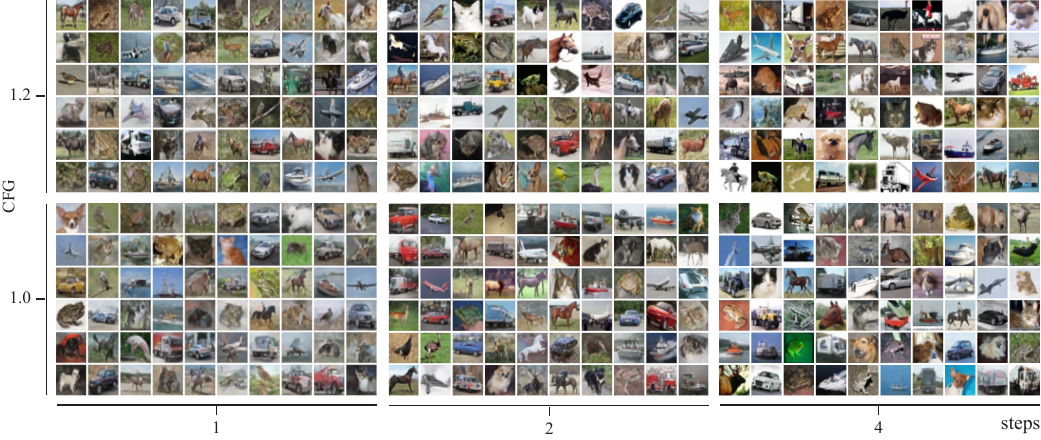}}
  \vskip -0.1in
  \caption{
    \textbf{Uncurated CIFAR-10 conditional generation} with different sampling steps and with/without CFG.
    Convincing images can be formed with a single forward pass, with reliability and gFID improving with up to 4 steps.
  }
  \label{fig:cifar}
  \vspace{-0.5em}
\end{figure*}

\begin{table}[t]
  \centering
  \scriptsize
  \scshape
  \caption{
    \textbf{Few-step generation results} (gFID $\downarrow$) on Animal-Faces (AF) and Oxford-Flowers (OF).
  }
  \label{tab:af_of}
  \setlength{\tabcolsep}{5pt} 
  \begin{tabular}{@{}lcccc@{}} 
    \toprule
    sphere-l model $\diagdown$ steps & 1       & 2     & 4     & 6     \\          
    \midrule
    of w/ cfg=1.6                    & 25.12   & 14.08 & 11.25 & 10.63 \\
    \cmidrule{1-5}
    of                               & 27.12   & 16.60 & 12.98 & 12.26 \\
    af                               & 21.70   & 19.29 & 18.23 & 17.97 \\
    \cmidrule{1-5}
    stylegan{\tiny{2}} \tinycite{style-gan-2}                          \\
    af - cat                         & \; 5.13 & -     & -     & -     \\
    af - dog                         & 19.37   & -     & -     & -     \\
    af - wild                        & \; 3.48 & -     & -     & -     \\
    \bottomrule
  \end{tabular}
  \vspace{-1.5em}
\end{table}

\cref{tab:cifar} presents both conditional and unconditional generation results.
For conditional generation, our method achieves strong results in both one-step and few-step generation, even without CFG.
For example, with less than $10$ sampling steps, sphere encoder yields gFID way below $2.0$ and IS above $10$.
For unconditional generation task, our method achieves better gFID and IS with less than $10$ sampling steps, a $100\times$ reduction in sampling steps, comparing to diffusion-based methods \cite{ddpm}.
\cref{fig:cifar} depicts qualitative results with different steps and CFG. Visually, our generation results without CFG look the same as those using CFG.
Appendix \ref{appsec:additional_cifar_results} provides quantitative results on CIFAR-10 with CFG.
Additionally, we discuss the memorization risk when training on small datasets like CIFAR-10 with extensive epochs in Appendix \ref{appsec:mem_risk}.

\subsection{Large Image Size}

We then evaluate our method with large image size ($256 \times 256$) on Oxford-Flowers \cite{oxford-flowers} ($8$K images), Animal-Faces \cite{starganv2} ($16$K images), and ImageNet \cite{imagenet} ($1.2$M images).

\textbf{Animal-Faces and Oxford-Flowers}.
We employ a ViT Large for encoder and decoder, \ie, Sphere-L, with a latent dimension of $L = 32\times32\times128$, training for $1000$ epochs.
Due to the low diversity of Animal-Faces ($3$ classes), we train unconditional models, while for Oxford-Flowers, we utilize conditional generation for the $102$ classes.
For evaluation, we adhere to the standard protocol of randomly sampling $50$K images, even though the training sets are relatively small.
\cref{tab:af_of} reports quantitative results on both datasets.
We report metrics only up to $6$ sampling steps, as performance saturates beyond this point.
Uncurated qualitative results are provided in \cref{fig:qualitative_results_afof_2steps,fig:qualitative_results_afof_4steps}.

\begin{figure*}[t]
  \vskip -0.1in
  \centering
  \centerline{\includegraphics[width=1.0\linewidth]{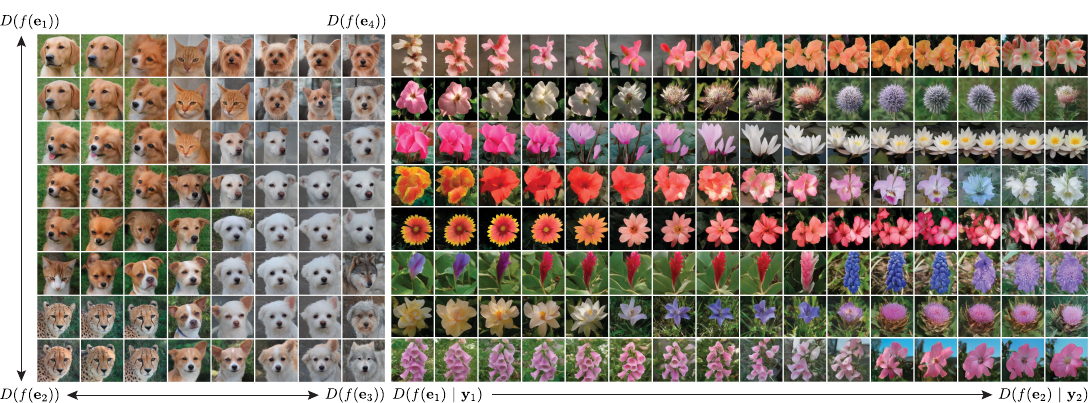}}
  \caption{
    \textbf{Latent interpolation} on Animal-Faces and Oxford-Flowers.
    Images are generated in 4 steps without CFG.
    (left) Interpolation in a 2D space that spans 4 synthetic images.
    (right) Each row interpolates between a random vector $\be$ on the sphere and a class conditional vector $\y$.
    Note that our model exhibits fast/sudden transitions between image classes rather than producing ``hybrid'' images that unrealistically merge properties of different object types.
    This property is necessary for a model to reliably convert random samples from the sphere into realistic images, as it makes the probability of observing a hybrid image small.
  }
  \label{fig:lerp}
  \vspace{-1.5em}
\end{figure*}

\textbf{ImageNet}.
We train class-conditional ViT-Large/-XLarge models on ImageNet for $800$ epochs, utilizing a latent dimension of $L = 32 \times 32 \times 64$. \cref{tab:imagenet} evaluates our approach alongside GANs, diffusion models, and other direct pixel-space generation frameworks.
At comparable parameter counts, our sphere encoder achieves competitive FID scores while requiring fewer sampling steps; its performance falls within the range of recent high-performing models, outperforming several established baselines.

\subsection{Lower FID scores?}

Because low FID scores do not always align with perceptual realism~\cite{stein2023exposing}, \cref{tab:imagenet} reports FIDs for our qualitatively strongest models, prioritizing visual quality over the optimization of a single numerical metric. Lower FID scores are possible with some tradeoffs.
For example, while training on CIFAR-10 for $10$K epochs can reduce FID to $0.94$, it does not yield a proportional gain in visual clarity.
On ImageNet, increasing sampling steps beyond $4$ improves the FID to $3.9$ and sharpens local edges, yet this numerical gain can introduce more abstract object structures.
This phenomenon -- where FID rewards local texture refinement even at the cost of global semantic coherence~\cite{jayasumana2024rethinking} — highlights a nuanced trade-off in generative modeling that warrants further investigation.

Still, our reported FID scores trail some recent high-performance generative models. We suspect this may be due to our use of pixel-space similarity losses, which likely contributes to the subtle edge blurring observed in our uncurated results (\cref{fig:qualitative_results_a,fig:qualitative_results_b}). In contrast, other recent works achieve high sharpness and low-FID through similarity metrics based purely on latent-space representations or multi-stage GAN losses -- a direction that should be evaluated in future work.

Finally, note that our training and conditioning methods do not rely on the discreteness of the ImageNet ontology. This generality opens the door for our methods to be transferred to the text-to-image setting.

\begin{table}
    \vspace{0.3em}
    \defcitealias{biggan-deep}{b21}
    \defcitealias{stylegan-xl}{s22}
    \defcitealias{gigagan}{k23}
    \defcitealias{add}{d21}
    \defcitealias{sid}{h23}
    \defcitealias{sid2}{h24}
    \defcitealias{jit}{l25}
    \defcitealias{jetformer}{t25}
    \defcitealias{fractal-mar}{l25}
    \centering
    \scriptsize
    \scshape
    \caption{
        \textbf{Few-step generation results} on ImageNet $256 \times 256$.
    }
    \label{tab:imagenet}
    \setlength{\tabcolsep}{4.5pt} 
    \begin{tabular}{@{}llrrrr@{}} 
        \toprule
         & method
         & params                                    & steps & gFID $\downarrow$ & IS $\uparrow$ \\
        \midrule
        \multirow{3}{*}{\rotatebox[origin=c]{90}{
                \tiny{gans}
            }}
         & biggan-deep \tinycitepalias{biggan-deep}
         & $56$M
         & 1                                         & 6.90  & 171.4                             \\
         & stylegan-xl \tinycitepalias{stylegan-xl}
         & $166$M
         & 1                                         & 2.30  & 265.1                             \\
         & gigagan \tinycitepalias{gigagan}
         & $569$M
         & 1                                         & 3.45  & 225.5                             \\
        \cmidrule{1-6}
        \multirow{6}{*}{\rotatebox[origin=c]{90}{
                \tiny{diffusions}
            }}
         & adm-g \tinycitepalias{add}
         & $554$M
         & 250                                       & 4.59  & 186.7                             \\
         & adm-g
         & $554$M
         & 25                                        & 5.44  & -                                 \\
         & sid \tinycitepalias{sid}
         & $2$B
         & 1000                                      & 2.44  & 256.3                             \\
         & sid{\tiny{2}} \tinycitepalias{sid2}
         & -
         & 512                                       & 1.38  & -                                 \\
         & jit-h \tinycitepalias{jit}
         & $953$M
         & 100                                       & 1.86  & 303.4                             \\
         & jit-g
         & $2$B
         & 100                                       & 1.82  & 292.6                             \\
        \cmidrule{1-6}
        \multirow{2}{*}{\rotatebox[origin=c]{90}{

            }}
         & jetformer \tinycitepalias{jetformer}
         & $2.8$B
         & -                                         & 6.64  & -                                 \\
         & fractalmar-h \tinycitepalias{fractal-mar}
         & $848$M                                    & 96    & 6.15              & 348.9         \\
        \cmidrule{1-6}
        \multirow{2}{*}{\rotatebox[origin=c]{90}{{
                        \tiny{ours}
                    }}}
         & sphere-l
         & $950$M
         & 4                                         & 4.76  & 301.8                             \\
         & sphere-xl
         & $1.3$B
         & 4                                         & 4.02  & 265.9                             \\
        \bottomrule
    \end{tabular}
    \vspace{-1.5em}
\end{table}

\section{Qualitative Experiments}
\label{sec:qualitative_properties}

\textbf{Latent Interpolation}.
In \cref{fig:lerp}, we interpolate between latent vectors on the Animal-Faces and Oxford-Flowers models to investigate the learned latent manifold.
On Animal-Faces, we randomly sample four noise vectors $\be$ in \cref{eq:generation} and visualize their corresponding images at the corners of the figure. We interpolate the latent space bilinearly to fill in the other images.
On Oxford-Flowers, we randomly sample a noise vector $\be$ and a class for each side of each row.
Since the model is class-conditional, we interpolate both input noise and class embeddings linearly as we move horizontally across each row.

We see that the model exhibits {\em fast transitions} between object types as we move through latent space.  For example, starting with the bottom-left image of a cheetah, we observe a sudden transition from cheetah to cat as we move vertically, and from cheetah to dog as we move horizontally. The model does not linger in a half-cheetah / half-dog state that is absent in the training data. These fast-transitions are required of a reliable generative model, as the probability of sampling an impossible/hybrid image should be minimized. This important property of the sphere encoder differentiates it from other latent models.  GANs, for example, tend to exhibit slow transitions, resulting in frequent production of distorted objects, \eg, Figure 8 and 9 in \citep{biggan-deep}.

\begin{figure}
  \vspace{-0.3em}
  \centering
  \centerline{\includegraphics[width=1.07\linewidth]{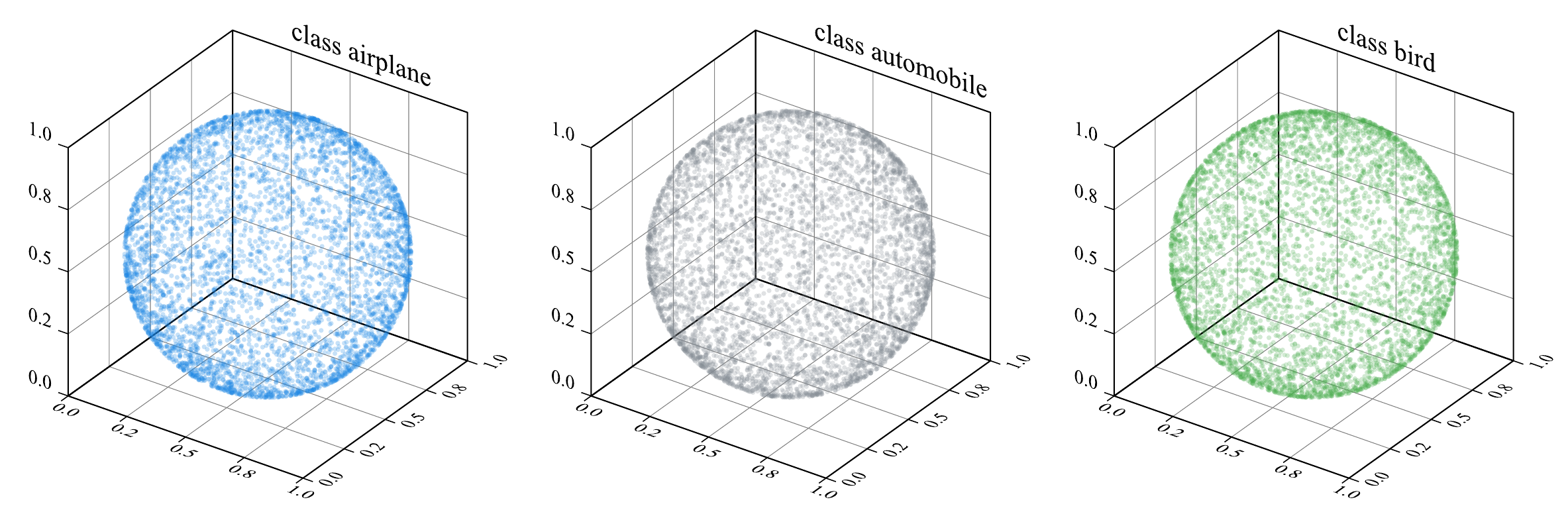}}
  \caption{
    \textbf{Latent space visualization} using random projection on CIFAR-10 training set.
    Each sphere shows the latent vectors of a different class.  The conditional latent distributions appear approximately uniform.
  }
  \label{fig:sphere_ball}
  \vspace{-2.0em}
\end{figure}

\begin{figure*}[b]
  \centering
  \centerline{\includegraphics[width=1.0\linewidth]{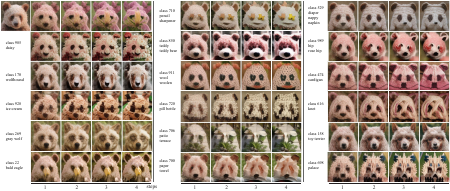}}
  \caption{
    \textbf{Conditional manipulation} via iterative encoding and decoding on ImageNet model. We demonstrate the model's expressivity using an out-of-domain input (a ``woolly panda'' top-left).
    Each row shows the result of conditioning the iterative process on different ImageNet classes without CFG.
    For example, the first row is conditioned on class 580 (greenhouse, nursery, glasshouse).
  }
  \label{fig:cond_manipulation}
\end{figure*}

\textbf{Conditional Uniformity}.
The latent distribution of our model should be uniform on the sphere.
For conditional models, the latent distribution must be {\em conditionally} uniform. To understand why, consider a conditional model with two classes, cat and dog. Suppose we have an unconditional encoder and a conditional decoder. It is likely that an unconditional encoder would structure latent space with cats in one region and dogs in another.  Even if the union of the two classes achieves uniform coverage, conditional generation may fail; we cannot reliably decode a dog from a random vector, as half the time this vector will represent a cat.

We avoid this pitfall by using a conditional encoder. In this case, our training objective naturally creates a latent distribution that is conditionally uniform, \ie, dogs alone uniformly cover the sphere, as do cats.  As a result, any uniformly sampled vector can be decoded to create the desired object.

To better understand the latent space learned by our method, we visualize it in \cref{fig:sphere_ball}.
We extract $\bv$ in \cref{eq:naive_spherify} for all CIFAR-10 $50$K training samples.
We project latents to $3$D space using a random Gaussian matrix, and then normalize each projected vector to have unit length.

We visualize the results separately for three random classes (other classes look similar).
We see that the latent space achieves conditional uniformity -- we get even coverage of the sphere for each class in isolation.

\begin{figure*}[t]
  \centering
  \centerline{\includegraphics[width=1.0\linewidth]{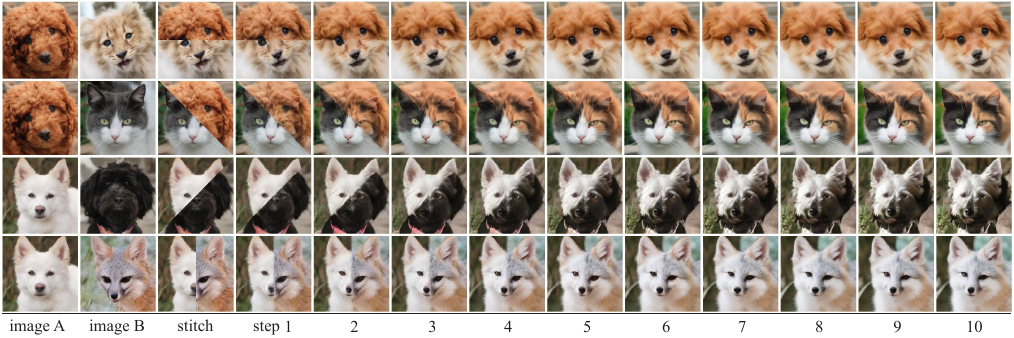}}
  \caption{
    \textbf{Image crossover} using the sphere encoder trained on Animal-Faces.
    A composite of two images (A and B) is iteratively processed through the encoder-decoder pipeline until it converges to a coherent sample on the learned image manifold.
  }
  \label{fig:edit_crossover}
  \vspace{-1.em}
\end{figure*}

\vspace{0.35em}
\section{Image Editing}
\label{sec:image_editing}

This section presents two {\it{training-free}} editing applications that leverage the expressivity and robustness of our latent space: simple semantic manipulation and image crossover.

\textbf{Conditional Manipulation}.
Given an out-of-domain image, \eg, a ``woolly panda'' in \cref{fig:cond_manipulation}, we can manipulate it by conditioning on different ImageNet classes.
We simply encode and decode the image with multiple steps, using the Sphere-L model (in \cref{tab:imagenet}) trained on ImageNet.
We set a fixed noise strength $r=1.0$ in \cref{eq:sigma_jitter} and $\gamma=0$ in \cref{eq:decay_schedule} across all steps, and do not apply CFG.

\cref{fig:cond_manipulation} demonstrates that a single step effectively captures the primary structure of the input while adapting its texture to align with the target class.
Subsequent iterations (\eg, 4-step generation) further refine these class-specific characteristics and textures, achieving semantic alignment while preserving the original image's structural integrity.

\textbf{Image Crossover}.
We further demonstrate the model's capability for ``image crossover'' by processing manually stitched composites of distinct source images (\cref{fig:edit_crossover}). This process similarly operates without CFG.
By repeatedly encoding and decoding the stitched composite (up to $10$ steps), the model naturally harmonizes the content and smooths boundary discontinuities. For these experiments, we set the noise magnitude $r=0.25$ (\cref{eq:sigma_jitter}) and apply a decay schedule (\cref{eq:decay_schedule}) with $\gamma=1$, which we found yields the most seamless blending.

This iterative refinement forces the manipulated image to converge toward a valid point on the learned spherical manifold. Notably, unlike diffusion models, \eg, Figure 12 in \cite{rectified-flow}, that require noise injection before projecting onto the image manifold, our encoder directly projects the stitched image into the latent space without adding noise (through the encoder).
This deterministic path preserves the semantic integrity of the original sources while ensuring a fluid, natural transition between them.

\section{Main Ablations}
\label{sec:ablations}

This section presents ablation studies on key design choices of our method.
Additional ablations regarding CFG strategies, noise distribution, uniform regularization, volume compression ratio, and model size are provided in Appendix \ref{appsec:additional_ablations}.

\textbf{Determining Noise Magnitude}.
In this section, we analyze the maximum noise magnitude $\sigma_{\text{max}}$ in \cref{eq:noisy_spherify} from a geometric perspective to understand its impact, and empirically determine its optimal value.
We begin with the noise-to-signal ratio (NSR) $\eta$ as the ratio of the expected noise magnitude to the signal magnitude in high-dimensional space:
\begin{align}
  \label{eq:nsr}
  \eta = \mathbb{E} \frac{\| \be \|}{\| \bv \|} = \frac{\sigma_{\text{max}} \sqrt{L}}{\sqrt{L}} = \sigma_{\text{max}}\eqdot
\end{align}
Because of the concentration of measure phenomenon, $\be$ is nearly orthogonal to $\bv$, \ie, $\bv^\top \be = 0$ when $L \to \infty$.
Thus, the angle $\alpha$ formed between $\bv + \sigma \cdot \be$ and $\bv$ satisfies:
\begin{align}
  \label{eq:alpha}
  \text{tan}(\alpha) \approx \frac{\| \be \|}{\| \bv \|} \approx  \sigma_{\text{max}} = \eta \eqcomma
\end{align}
which gives $\sigma_{\text{max}}$ an interpretable geometric meaning.
The noise magnitude $\sigma_{\text{max}}$ can be equivalently expressed by either the angle $\alpha$ or the NSR $\eta$.

We build a conditional encoder and decoder using classic ViT with Base size ($12$ layers), and train them on ImageNet for $200$ epochs.
The latent dimension is $L = 16\times16\times256$.
We vary $\alpha$ from $45^\circ$ to $88^\circ$, corresponding to $\sigma_{\text{max}} = \tan(\alpha)$ from $1$ to $28.6$.

\begin{figure}[t]
  \vspace{-0.25em}
  \centering
  \centerline{\includegraphics[width=1.0\linewidth]{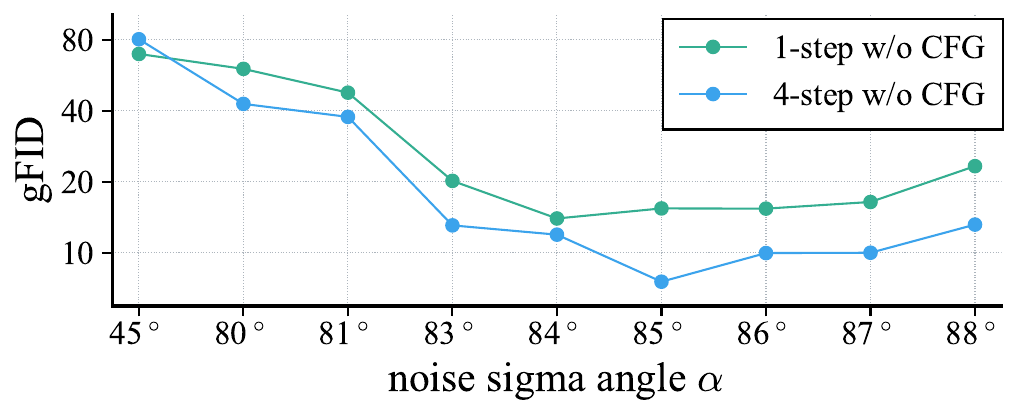}}
  \vspace{-0.25em}
  \caption{
    \textbf{Quantitative impact of the angle $\alpha$} on ImageNet.
  }
  \label{fig:sigma_angle_curv}
  \vspace{0.5em}
  \centerline{\includegraphics[width=1.0\linewidth]{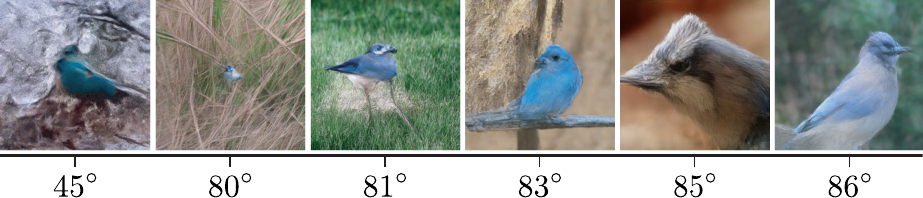}}
  \caption{
    \textbf{Qualitative impact of the angle $\alpha$} on ImageNet with 4-step generation.
  }
  \label{fig:sigma_angle_imgs}
  \vspace{-1.0em}
\end{figure}

Our first ablations are done to select the noise level.
The NSR $\eta$ guides the difficulty of the decoder's task.
The decoder aims to generate the same image from $\bv_\text{NOISY}$ as from the clean $\bv$.
When $\alpha \leq 45^\circ$ (equivalently $\eta \leq 1$), the noise does not overwhelm $\bv$ and the latent clouds generated by each training point are well separated. The decoder reconstructs images well, but the noisy latents fail to cover the sphere. In this case, generation from random sampling using \cref{eq:generation} fails, with a high gFID in \cref{fig:sigma_angle_curv}.

As $\alpha \to 90^\circ$, the latent representations of images are forced apart and the decoder starts to generate images from random latents. The gFID drops significantly in \cref{fig:sigma_angle_curv}.
\cref{fig:sigma_angle_imgs} shows some sampled images with different $\alpha$, demonstrating that a lower $\alpha$ leads to abstract and blurry images, while a higher $\alpha$ produces more realistic details.

Once we dial in $\alpha$, we find that we can fix it for various latent dimensions $L$ because of the dimension-invariant property of angle $\alpha$ in \cref{eq:nsr}.
However, we found that the optimal $\alpha$ varies slightly with image size.
For small image size, \eg, $32\times32$ on CIFAR-10, we found $\alpha = 80^\circ$ works best.
For large image size, \eg, $256\times256$ on ImageNet, we found $\alpha = 85^\circ$ works best.

\begin{table}
  \centering
  \scriptsize
  \scshape
  \caption{
    \textbf{Ablation of loss terms} on ImageNet.
    Loss terms are added incrementally from top to bottom.
  }
  \label{tab:loss}
  \setlength{\tabcolsep}{5pt} 
  \begin{tabular}{@{}lccrcrc@{}} 
    \toprule
                                      &
                                      &
                                      & \multicolumn{2}{c}{w/o cfg}
                                      & \multicolumn{2}{c}{w/ cfg $=1.6$}                 \\
    \cmidrule(r){4-5} \cmidrule(l){6-7}
    loss $\cL$
                                      & steps
                                      & rFID $\downarrow$
                                      & gFID $\downarrow$                 & IS $\uparrow$
                                      & gFID $\downarrow$                 & IS $\uparrow$ \\
    \midrule
    $\cL_{\text{pix-recon}}$          & 1                                 & 1.70
                                      & 25.35                             & 168.8
                                      & 20.68                             & 204.4         \\
                                      & 4                                 & -
                                      & 13.58                             & 162.9
                                      & 11.76                             & 231.7         \\
    \cmidrule{1-7}
    {\tiny{+}} $\cL_{\text{pix-con}}$ & 1                                 & 0.89
                                      & 17.19                             & 183.8
                                      & 14.74                             & 218.3         \\
                                      & 4                                 & -
                                      & 10.12                             & 189.9
                                      & 10.47                             & 247.1         \\
    \cmidrule{1-7}
    {\tiny{+}} $\cL_{\text{lat-con}}$ & 1                                 & 1.32
                                      & 15.40                             & 188.3
                                      & 13.14                             & 225.1         \\
                                      & 4                                 & -
                                      & 7.53                              & 176.4
                                      & 7.28                              & 243.1         \\
    \bottomrule
  \end{tabular}
  \vspace{-1.em}
\end{table}

\textbf{Training Loss}.
We evaluate the individual contribution of each proposed loss term in \cref{sec:losses} to generation quality.
Adopting the same ImageNet setup as the previous ablation, \cref{tab:loss} reports the results of adding each term incrementally.

Starting with only the pixel reconstruction loss $\cL_{\text{pix-recon}}$, we observe the model can generate images, but the quality is suboptimal with a serious ``waffle'' artifact.
Adding the pixel consistency loss significantly improves generation quality by removing the artifact, as it encourages the decoder to produce consistent images from perturbed latents.

\begin{figure}[t]
  \centering
  \centerline{\includegraphics[width=1.0\linewidth]{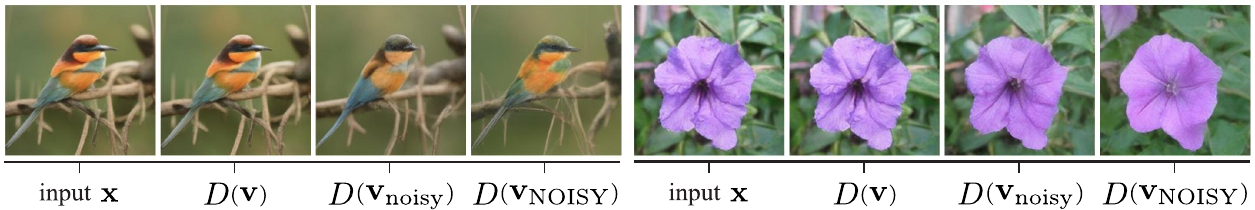}}
  \caption{
    \textbf{Consistency optimization path} from noisy latent to clean latent on the sphere, pushing the decoder to generate consistent and diverse images from right to left.
  }
  \label{fig:consistency_optim_path}
  \vspace{-0.5em}
\end{figure}

Including the latent consistency loss yields the best performance, demonstrating its effectiveness in guiding the encoder-decoder pair to learn a coherent latent manifold.
\cref{fig:consistency_optim_path} visualizes the optimization path from noisy latent to clean latent on the sphere for those two consistency losses during training.
Overall, each loss contributes positively to the model's ability to generate high-quality images.

\textbf{Latent Spatial Resolution}.
In these ablations, we keep the latent dimension constant and vary the latent spatial resolution by adjusting channel depth $d$ of latent dim $L$.
\cref{tab:latent_spatial_resolution} presents the results on ImageNet, suggesting that a higher latent spatial resolution with volume compression ratio $3.0$ works best on ImageNet, \ie, $L = 32^2 \times 64$.
On CIFAR-10, Animal-Faces, and Oxford-Flowers, we also observed that a higher latent spatial resolution yields better generation quality, but with a lower compression ratio $1.5$.
The finalized latent dimensions are detailed in Appendix \ref{appsec:hyperparams}.

\begin{table}[t]
  \centering
  \scriptsize
  \scshape
  \caption{
    \textbf{Ablation on latent spatial resolution} with two optimal compression ratios on ImageNet.
  }
  \label{tab:latent_spatial_resolution}
  \setlength{\tabcolsep}{3.8pt} 
  \begin{tabular}{@{}lccrrrr@{}} 
    \toprule
     &                                   &
     & \multicolumn{2}{c}{w/o cfg}
     & \multicolumn{2}{c}{w/ cfg $=1.2$}                     \\
    \cmidrule(r){4-5} \cmidrule(l){6-7}
    latent $L$
     & steps                             & rFID $\downarrow$
     & gFID$\downarrow$                  & IS $\uparrow$
     & gFID $\downarrow$                 & IS $\uparrow$     \\
    \midrule
    \multicolumn{7}{c}{volume compression ratio $=1.5$}      \\
    \cmidrule{1-7}
    $16^2 \times 512$
     & 1                                 & 0.64
     & 16.06                             & 169.8
     & 14.25                             & 187.6             \\
     & 4                                 & -
     & 10.03                             & 153.0
     & 8.71                              & 180.8             \\
    \cmidrule{1-7}
    $32^2\times128$
     & 1                                 & 0.60
     & 20.63                             & 145.1
     & 17.69                             & 164.9             \\
     & 4                                 & -
     & 15.45                             & 127.0
     & 12.75                             & 154.6             \\
    \midrule
    \multicolumn{7}{c}{volume compression ratio $=3.0$}      \\
    \cmidrule{1-7}
    $16^2\times256$
     & 1                                 & 0.74
     & 16.93                             & 206.3
     & 15.39                             & 227.7             \\
     & 4                                 & -
     & 8.30                              & 202.9
     & 7.85                              & 232.3             \\
    \cmidrule{1-7}
    $32^2\times64$
     & 1                                 & 0.61
     & 16.80                             & 227.8
     & 15.33                             & 267.0             \\
     & 4                                 & -
     & 7.65                              & 219.4
     & 7.48                              & 252.4             \\
    \bottomrule
  \end{tabular}
  \vspace{-1.em}
\end{table}
\begin{table}[t]
  \vspace{0.25em}
  \centering
  \scriptsize
  \scshape
  \caption{
    \textbf{Ablation on sampling schemes} with few-step generation and no CFG.
    Results are reported with gFID $\downarrow$ on ImageNet.
  }
  \label{tab:sampling_schedule}
  \setlength{\tabcolsep}{5pt} 
  \begin{tabular}{@{}cccccccc@{}} 
    \toprule
    $\gamma$ & share $\be$ & 1     & 2    & 4    & 6       & 8       & 10      \\
    \midrule
    0        & -           & 16.57 & 7.99 & 7.68 & \; 7.49 & \; 7.61 & \; 7.71 \\
             & \checkmark  & 16.66 & 7.97 & 5.99 & \; 7.78 & 10.11   & 12.25   \\
    \cmidrule{1-8}
    1        & -           & 16.73 & 8.12 & 9.02 & 12.17   & 15.08   & 18.09   \\
             & \checkmark  & 16.81 & 8.02 & 8.53 & 12.32   & 17.50   & 22.78   \\
    \bottomrule
  \end{tabular}
  \vspace{-1.em}
\end{table}

\textbf{Sampling Schemes}.
In \cref{alg:code}, the few-step generation involves adding noise in spherifying at each step following \cref{eq:noisy_spherify}.
We investigate two key aspects of this noise injection mechanism: strength schedule and sampling schedule.
First, regarding the noise strength controlled by $r$ in \cref{eq:sigma_jitter}, we compare two cases:
(a) a fixed $r = 1.0$ for all steps,
and (b) a decaying schedule, where $r$ decreases from $1.0$ to a minimum value.
Second, we also have two options for sampling the noise vector $\be$ in \cref{eq:noisy_spherify}:
(a) sampling independent noise $\be$ for each step,
versus (b) sharing the same noise $\be$ across all steps.
We qualitatively and quantitatively evaluate both aspects without CFG.

We simply decay the noise strength $r$ with a linear schedule:
\begin{align}
  \label{eq:decay_schedule}
  r = \left(1 - \frac{t-1}{T-1}\right)^\gamma \eqcomma
\end{align}
where $t$ is the current step from $2$ to $T$ in the loop, and decay rate $\gamma=1$.
The $t=1$ is the first step of decoder forward pass, which is not involved with spherifying process.
When $\gamma = 0$, it corresponds to the fixed schedule, \ie, $r = 1.0$.

\cref{tab:sampling_schedule} presents the results on ImageNet with those four sampling schemes.
The fixed schedule ($\gamma=0$) outperforms the decaying schedule ($\gamma=1$) across all sampling steps.
In addition, sharing the same noise $\be$ across all steps consistently yields better results than using independent noise for each step.
We hypothesize that sharing the same noise helps maintain a consistent direction during the optimization path on the sphere, leading to more stable and effective generation.
Overall, the best sampling scheme is using a fixed noise strength with shared noise across all steps.

\begin{figure}[t]
  \centering
  \centerline{\includegraphics[width=1.0\linewidth]{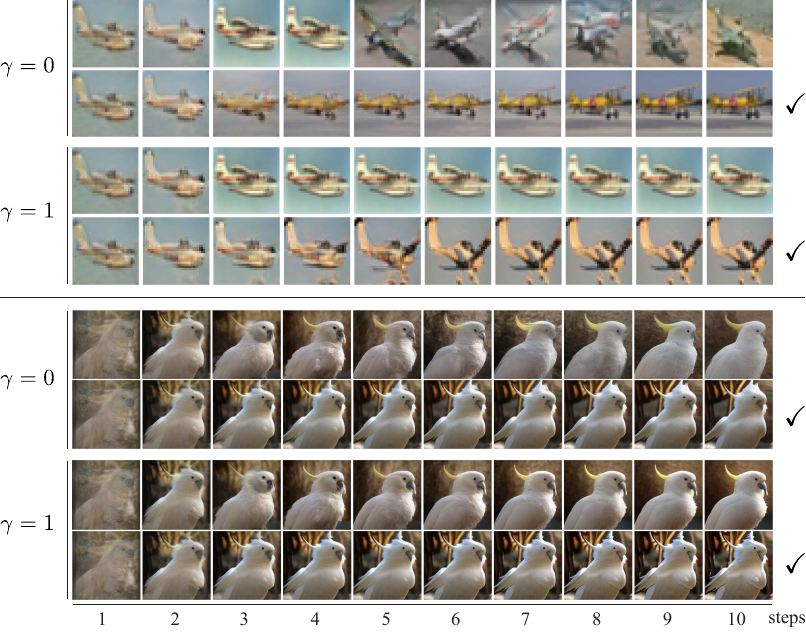}}
  \caption{
    \textbf{Qualitative impact of sampling schemes} on generation without CFG for CIFAR-10 (top) and ImageNet (bottom).
    The $\checkmark$ indicates sharing the same noise $\be$ across steps.
    Shared noise with $\gamma=1$ yields superior coherence and a sharp ``paper art'' aesthetic on ImageNet as sampling steps increase.
  }
  \label{fig:sampling_schedule}
  \vspace{-0.75em}
\end{figure}

\cref{fig:sampling_schedule} shows generated images on both CIFAR-10 and ImageNet under different sampling schemes.
Visual inspection confirms that sharing the same noise across steps yields significantly more coherent and higher-quality results than using independent noise.
Notably, on ImageNet, we observe that sharing noise with decay schedule $\gamma=1$ produces exceptionally sharp images as the number of sampling steps increases.
For example, with $10$ steps, the images exhibit a distinct, hyper-sharp aesthetic reminiscent of paper art.

\textbf{Achieving a Uniform Distribution}.
Our method does not employ explicit regularization to achieve a uniform latent distribution on the sphere. Comprehensive ablation studies in Appendix \ref{appsec:explicit_dist_reg} demonstrate that such regularization is unnecessary, as our method naturally achieves the desired latent properties.

\section{Related Work}

\textbf{Spherical latent space} has been explored primarily through variational inference, using priors such as the von Mises-Fisher distribution \cite{z2018spherical,sae,zde2020power,sae-kl}.
However, these approaches inherit limitations from VAEs, including significant posterior-prior mismatch.
These issues are compounded in high-dimensional latent spaces, where sampling relies on intricate reparameterization or rejection sampling techniques.
Although \cite{zzhao2019latent} drew inspiration from StyleGAN \cite{style-gan-1,style-gan-2,progressive-gan} to enable direct sampling in a high-dimensional unit spherical space via simple normalization, their experiments were limited to toy datasets like MNIST.
Since VAEs can already perform direct sampling on MNIST \cite{diagnosing-ae,taming-vae}, the advantages of spherical latent spaces in this context remain unclear.
Crucially, the potential of high-dimensional spherical latent spaces \cite{rjf} for image generation remains significantly underexplored.

\textbf{Few-step generation} has been extensively studied in both GANs and diffusion models.
GANs \cite{gan} are inherently created for one-step generation.
While approaches like ProgressiveGAN \cite{progressive-gan} introduce multi-stage refinement, these primarily serve as training strategies \cite{lapgan,stackgan,biggan-deep} rather than altering the fundamental single-pass inference process.
Conversely, diffusion models rely on an iterative generation process, typically requiring hundreds to thousands of steps.
Although distillation \cite{one-step-shortcut-models,one-step-dist-matching,one-step-em-distillation,one-step-progressive-distillation,mean-flows-1,adversarial-diffusion-distillation} and consistency techniques \cite{consistency-models,consistency-models-easy,consistency-flow-matching,improved-consistency-models} have been proposed to accelerate this to a few steps, their core insight aims to approximate the original diffusion trajectory.

\textbf{Pixel-space generation} is common for GANs \cite{style-gan-2,gigagan,stylegan-xl,biggan-deep,biggans} but challenging for diffusion models due to the high dimensionality of pixel space \cite{sparse-transformers,pixelcnn,i-gpt,perceiver-ar,megabyte}.
Recent advances based on diffusion mechanisms have made strides in pixel-space generation \cite{fractal-mar,jit,pixel-dit}.
Our sphere encoder diverges fundamentally from both paradigms.
Its few-step mechanism iteratively traverses between latent and pixel spaces, grounded in a spherical latent space.

\textbf{Signal processing.}  Concepts from this work take inspiration from sphere encoders/decoders in wireless networking that distribute codewords uniformly across a sphere \citep{studer2008soft,studer2010soft}.

\section{Conclusion}
The Sphere encoder is a novel generative framework that enables few-step, high-fidelity image synthesis. Our key observation is that the distribution of latents can be tightly controlled on a uniform sphere, enabling the training of an autoencoder that can be directly sampled.

\begin{figure}[t]
  \centering
  \centerline{\includegraphics[width=1.0\linewidth]{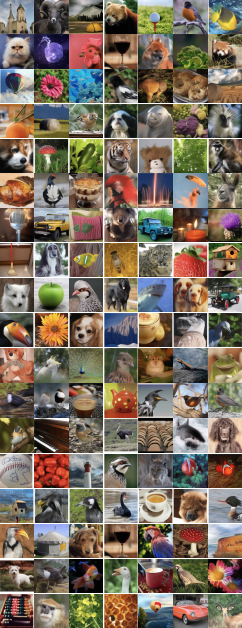}}
  \caption{
    \textbf{Randomly selected images generated by the Sphere encoder} for ImageNet ($256\times256$).
    Results are generated using Sphere-XL with 4-step sampling and CFG $=1.4$.
  }
  \label{fig:rs}
\end{figure}

This work is intended to be a proof-of-concept for direct conditional and unconditional generation from an autoencoder, and we suspect this first implementation is far from optimal.
To illustrate the average generation quality, \cref{fig:rs} shows randomly selected ImageNet results, with comprehensive uncurated examples provided in \cref{fig:qualitative_results_a,fig:qualitative_results_b}.

One drawback of our approach is that parameters must be allocated for both the encoder and decoder, and two passes are needed through the encoder during training, \ie, one for latent encoding and another for consistency loss.  Interestingly, improvements that enable single-pass generation for more complex distributions would eliminate the need for the encoder at inference time, and possibly at training time as well.  We think there are many avenues for future research that could unlock this and other capabilities, \eg, text-to-image generation.

\section*{Acknowledgements}
We would like to thank
\href{https://scholar.google.com/citations?user=wFduC9EAAAAJ&hl=en}{\textcolor{black}{Tan Wang}},
\href{https://scholar.google.com/citations?user=XYxv5HIAAAAJ&hl=en}{\textcolor{black}{Chenyang Zhang}},
\href{https://scholar.google.com/citations?user=wKjXhH8AAAAJ&hl=en}{\textcolor{black}{Tian Xie}},
\href{https://scholar.google.com/citations?user=yFMX138AAAAJ&hl=en}{\textcolor{black}{Wei Liu}},
\href{https://scholar.google.com/citations?user=dgN8vtwAAAAJ&hl=en}{\textcolor{black}{Felix Juefei Xu}},
and \href{https://scholar.google.com/citations?user=NWPDSEsAAAAJ&hl=en}{\textcolor{black}{Andrej Risteski}}
for their valuable discussions and feedback.

\clearpage
\appendix


\begin{figure*}[t]
  \centering
  \centerline{\includegraphics[width=1.0\linewidth]{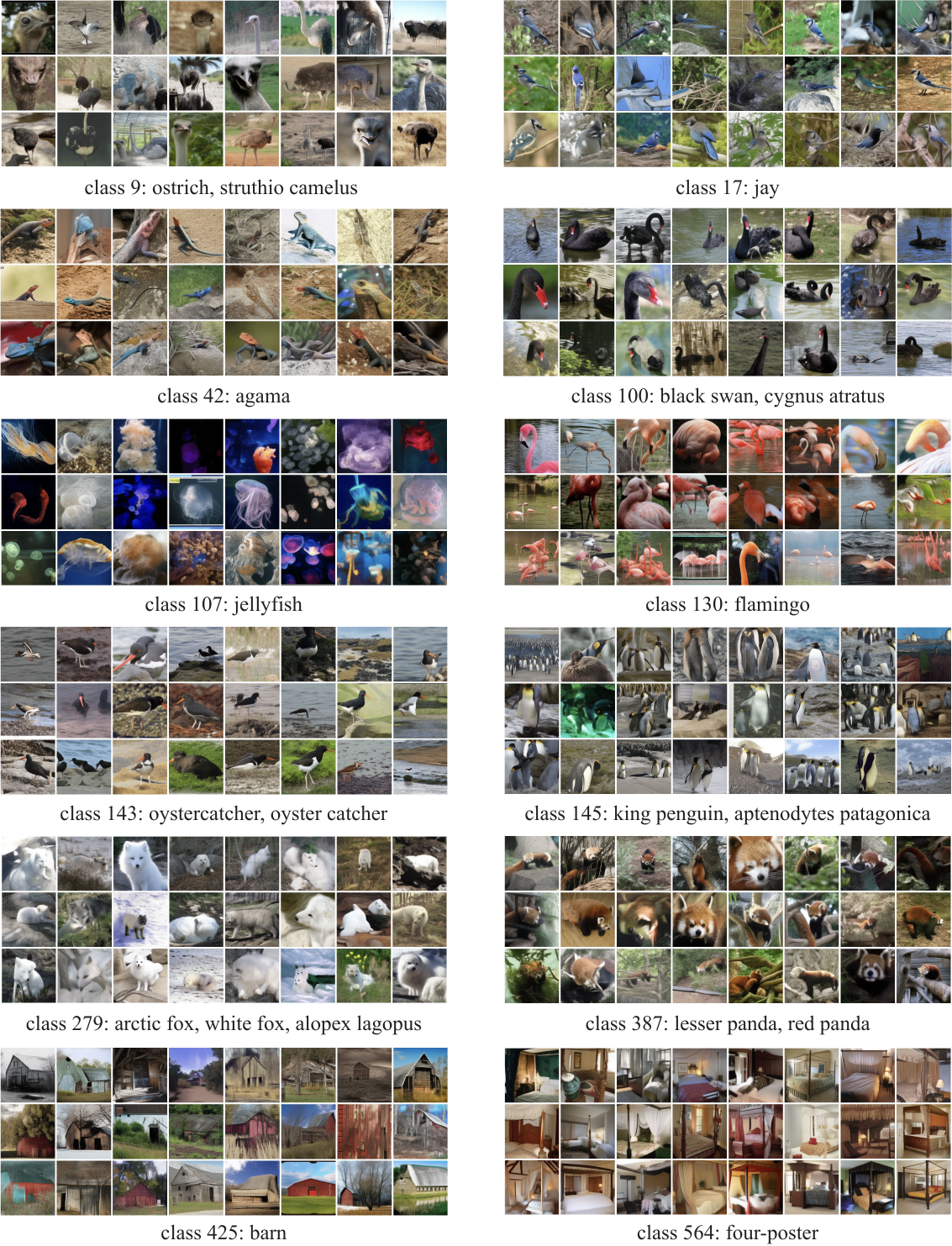}}
  \caption{
    \textbf{Uncurated results} on ImageNet ($256\times256$).
    Results are generated using Sphere-XL with 4-step sampling and CFG $=1.4$.
  }
  \label{fig:qualitative_results_a}
\end{figure*}

\clearpage
\begin{figure*}[t]
  \centering
  \centerline{\includegraphics[width=1.0\linewidth]{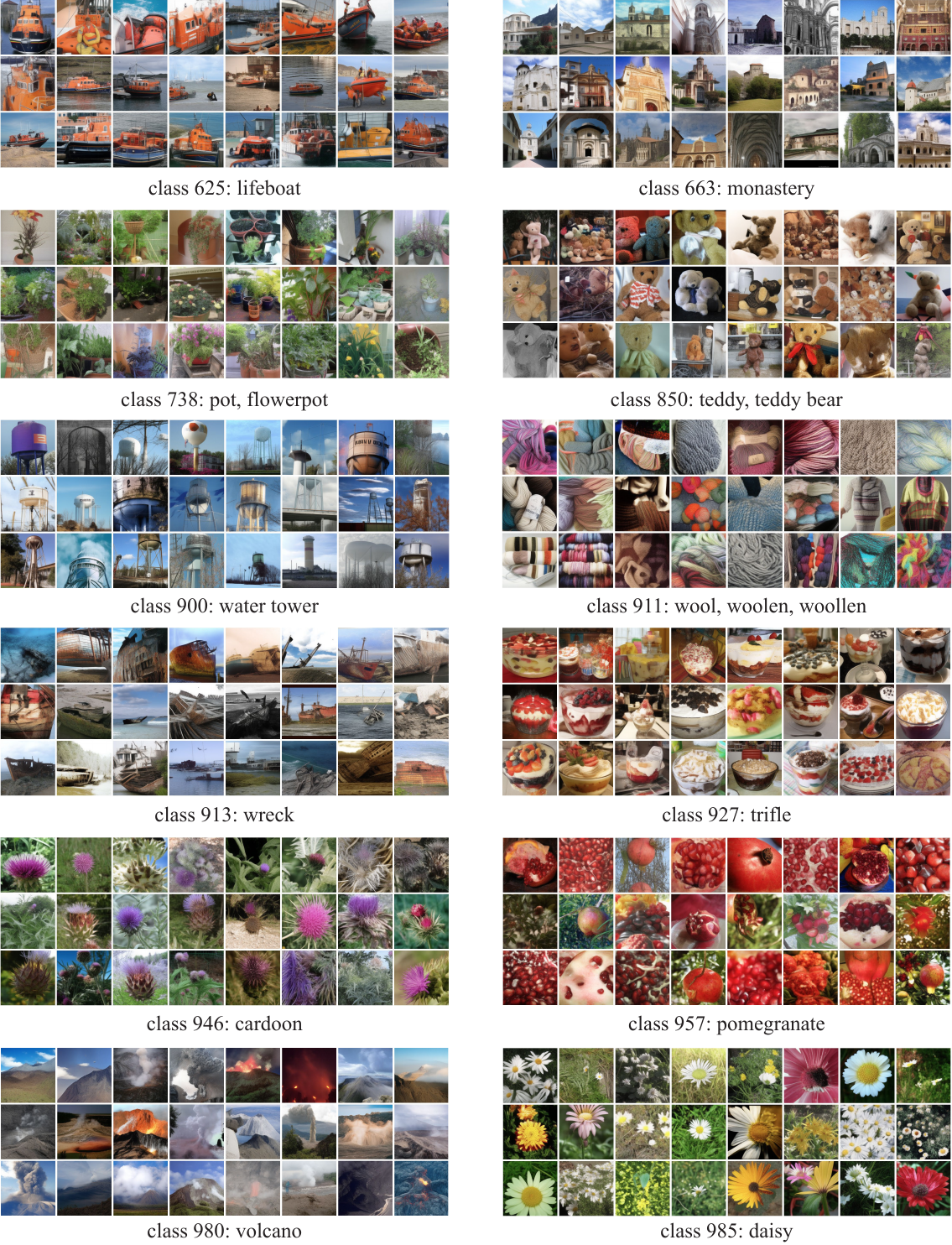}}
  \caption{
    \textbf{Uncurated results} on ImageNet ($256\times256$).
    Results are generated using Sphere-XL with 4-step sampling and CFG $=1.4$.
  }
  \label{fig:qualitative_results_b}
\end{figure*}

\clearpage
\begin{figure*}[t]
  \centering
  \centerline{\includegraphics[width=1.0\linewidth]{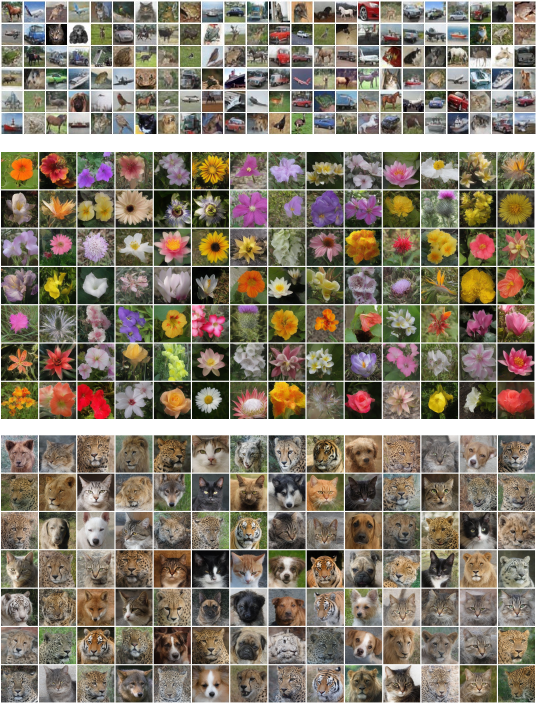}}
  \caption{
    \textbf{Uncurated qualitative results} on CIFAR-10 ($32\times32$), Oxford-Flowers and Animal-Faces ($256\times256$).
    Results are 2-step generation without CFG.
  }
  \label{fig:qualitative_results_afof_2steps}
\end{figure*}

\clearpage
\begin{figure*}[t]
  \centering
  \centerline{\includegraphics[width=1.0\linewidth]{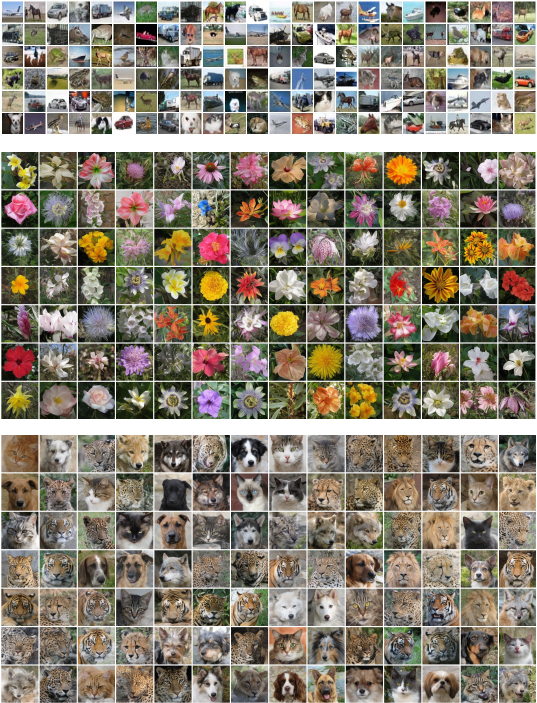}}
  \caption{
    \textbf{Uncurated qualitative results} on CIFAR-10 ($32\times32$), Oxford-Flowers and Animal-Faces ($256\times256$).
    Results are 4-step generation without CFG.
  }
  \label{fig:qualitative_results_afof_4steps}
\end{figure*}

\clearpage
\section{Additional Results on CIFAR-10}
\label{appsec:additional_cifar_results}

\cref{tab:cifar_with_cfg} shows that using CFG $=1.2$ yields a slight improvement over the baseline in \cref{tab:cifar}.
For example, with 6-step sampling, gFID decreases from $1.65$ (no CFG) to $1.41$ (with CFG), and IS increases from $10.7$ to $10.8$.

\section{Memorization Risk on CIFAR-10}
\label{appsec:mem_risk}

\begin{figure}[h]
  \centering
  \centerline{\includegraphics[width=1.0\linewidth]{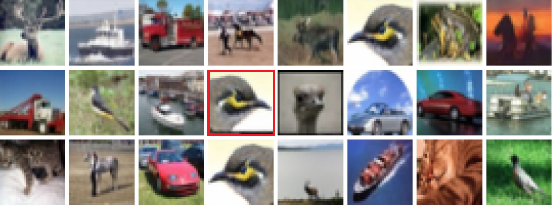}}
  \caption{
    \textbf{Memorization risk} with longer $10$K training epochs on CIFAR-10.
    Each row is a different sampling run showing near-duplicate birds of the real training image (in red box).
  }
  \label{fig:memorization}
\end{figure}

In the case of training longer on CIFAR-10, \eg, around $10$K epochs following common practice of diffusion models \cite{consistency-models,mean-flows-1}, we found our model sometimes generates near-duplicate samples.
\cref{fig:memorization} presents near-duplicate samples, \ie, the bird, from different sampling runs.
This phenomenon indicates our model may memorize some training samples when trained excessively to overfit the small-scale data distribution.
However, our model generates the flipped version of the bird, which is not exactly the same as the real image, suggesting that the model does not simply copy the training image but rather learns a transformation of it.
This aligns with known memorization issue in diffusion models \cite{diffusion-copying,detecting-memorization-diffusion}.
This longer training improves generation quality further, as shown in \cref{tab:cifar_long}, \eg, gFID $=0.94$ and IS $=11.1$ with 10-step sampling with CFG.

\section{Additional Ablations}
\label{appsec:additional_ablations}

\subsection{CFG Position}
\label{appsec:cfg_position}

In \cref{alg:code}, we have three options to apply CFG:
(1) only in the pixel space after decoding;
(2) only in the latent space after encoding;
(3) in both latent and pixel spaces, termed as ``combo''.
Since the combo option applies CFG twice, we halve the CFG scale for each application to keep the overall strength consistent.
For example, if the overall CFG scale is $s$, we use $s^{1/2}$ for each position.
\cref{tab:cfg} presents the results on ImageNet with different CFG positions and scales.
We observe that applying CFG in the pixel space consistently outperforms applying it in the latent space.
The combo option with CFG $=1.6$ achieves the best results for 4-step sampling on ImageNet.
Unless otherwise specified, we apply combo for all experiments.

\begin{table}
    \centering
    \scriptsize
    \scshape
    \caption{
        \textbf{Few-step conditional generation results} on CIFAR-10 with longer $10$K training epochs.
    }
    \label{tab:cifar_long}
    \setlength{\tabcolsep}{5pt} 
    \begin{tabular}{@{}ccrrr@{}} 
        \toprule
        method   & steps & rFID $\downarrow$ & gFID $\downarrow$ & IS $\uparrow$ \\
        \midrule
        \multicolumn{5}{c}{without CFG}                                          \\
        \cmidrule{1-5}
        sphere-l & 1     & 0.68              & 16.02             & 9.5           \\
                 & 2     & -                 & 4.92              & 10.3          \\
                 & 4     & -                 & 1.24              & 10.7          \\
                 & 6     & -                 & 1.01              & 10.9          \\
                 & 8     & -                 & 1.01              & 10.8          \\
                 & 10    & -                 & 1.00              & 10.9          \\
        \midrule
        \multicolumn{5}{c}{with CFG = 1.2}                                       \\
        \cmidrule{1-5}
        sphere-l & 1     & -                 & 16.57             & 9.5           \\
                 & 2     & -                 & 5.17              & 10.4          \\
                 & 4     & -                 & 1.20              & 10.8          \\
                 & 6     & -                 & 0.96              & 10.9          \\
                 & 8     & -                 & 0.94              & 10.9          \\
                 & 10    & -                 & 0.94              & 11.1          \\
        \bottomrule
    \end{tabular}
\end{table}

\begin{table}
  \centering
  \scriptsize
  \scshape
  \caption{
    \textbf{Few-step generation results} on CIFAR-10 with CFG.
  }
  \label{tab:cifar_with_cfg}
  \setlength{\tabcolsep}{5pt} 
  \begin{tabular}{@{}lccrrr@{}} 
    \toprule
    method   & cfg & steps & rFID $\downarrow$ & gFID $\downarrow$ & IS $\uparrow$ \\
    \midrule
    sphere-l & 1.2 & 1     & 0.59              & 18.82             & 9.1           \\
             &     & 2     & -                 & 8.41              & 10.0          \\
             &     & 4     & -                 & 2.53              & 10.6          \\
             &     & 6     & -                 & 1.41              & 10.8          \\
             &     & 8     & -                 & 1.18              & 10.8          \\
             &     & 10    & -                 & 1.12              & 10.9          \\
    \bottomrule
  \end{tabular}
\end{table}

\begin{table*}[t]
  \centering
  \scriptsize
  \scshape
  \caption{
    \textbf{Ablation on CFG position}.
    Results are reported with gFID $\downarrow$ on ImageNet.
    The sampling scheme is the fixed strength with sharing noise $\be$ in spheriying process across all steps.
  }
  \label{tab:cfg}
  \setlength{\tabcolsep}{5pt} 
  \begin{tabular}{@{}ccccccccccccccccccccc@{}} 
    \toprule
     & \multicolumn{5}{c}{steps w/ cfg $=1.0$}
     & \multicolumn{5}{c}{steps w/ cfg $=1.2$}
     & \multicolumn{5}{c}{steps w/ cfg $=1.4$}
     & \multicolumn{5}{c}{steps w/ cfg $=1.6$}                                             \\
    \cmidrule(lr){2-6}
    \cmidrule(lr){7-11}
    \cmidrule(lr){12-16}
    \cmidrule(l){17-21}
    cfg position
     & 1                                       & 2    & 4            & 6            & 8
     & 1                                       & 2    & 4            & 6            & 8
     & 1                                       & 2    & 4            & 6            & 8
     & 1                                       & 2    & 4            & 6            & 8    \\
    \midrule
    enc
     & 16.7                                    & 7.9  & 6.0          & 7.9          & 10.0
     & 16.7                                    & 8.0  & 6.1          & 7.8          & 10.3
     & 16.8                                    & 7.9  & 6.1          & 7.9          & 10.5
     & 16.8                                    & 7.9  & 6.11         & 8.1          & 10.7 \\
    dec
     & 16.7                                    & 8.0  & 5.9          & 7.8          & 10.1
     & 15.0                                    & 8.3  & \textbf{4.7} & 5.1          & 6.2
     & 15.4                                    & 9.3  & 5.0          & 4.5          & 5.0
     & 16.4                                    & 10.2 & 5.7          & \textbf{4.7} & 4.8  \\
    combo
     & 16.7                                    & 7.9  & 6.0          & 7.8          & 10.2
     & 15.3                                    & 8.0  & 5.1          & 6.1          & 7.9
     & 15.1                                    & 8.2  & \textbf{4.7} & 5.2          & 6.5
     & 15.0                                    & 8.6  & \textbf{4.7} & \textbf{4.7} & 5.7  \\
    \bottomrule
  \end{tabular}
  \vskip -0.1in
\end{table*}

\subsection{Dialing in the Noise Distribution}
\label{appsec:sampling_angle_range}
Because the noise magnitude has a strong impact on image quality, we dial this in by considering a few more sophisticated sampling strategies for the noise magnitude. After determining the optimal maximum angle $\alpha$ in the previous section, we further explore whether mixing a range of larger angles during training can enhance generation quality.
We conduct experiments on CIFAR-10 with latent size $L = 16\times16\times8$.
We train the model for $2000$ epochs.

In \cref{tab:noise_sigma_mix_hard_case}, we compare three settings:
(1) first row: a base case where angles are sampled uniformly from $[0^\circ, 80^\circ]$;
(2) second row: the base case mixing with larger angles from $[80^\circ, 85^\circ]$ with $0.1$ probability;
and (3) third row: the base case mixing with even larger angles from $[80^\circ, 89^\circ]$ with $0.1$ probability.

\begin{table}[h]
  \centering
  \scriptsize
  \scshape
  \caption{
    \textbf{Impact of mixing a range of larger angles} during training on CIFAR-10.
    All settings yield rFID scores in the range of $0.46 \text{--} 0.48$.
  }
  \label{tab:noise_sigma_mix_hard_case}
  \setlength{\tabcolsep}{5pt} 
  \begin{tabular}{@{}cccrrrr@{}} 
    \toprule
                      &                                   &
                      & \multicolumn{2}{c}{w/o cfg}
                      & \multicolumn{2}{c}{w/ cfg $=1.6$}                 \\
    \cmidrule(lr){4-5} \cmidrule(l){6-7}
    base max $\alpha$ & mix range                         & steps
                      & gFID $\downarrow$                 & IS $\uparrow$
                      & gFID $\downarrow$                 & IS $\uparrow$ \\
    \midrule
    $80^\circ$        & -                                 & 1
                      & 21.17                             & 8.8
                      & 18.93                             & 9.3           \\
                      & -                                 & 4
                      & 8.48                              & 10.0
                      & 7.47                              & 10.1          \\
    \cmidrule{1-7}
    $80^\circ$        & $[80^\circ, 85^\circ]$            & 1
                      & 19.82                             & 8.7
                      & 18.39                             & 9.1           \\
                      &                                   & 4
                      & 7.98                              & 10.1
                      & 7.05                              & 10.2          \\
    \cmidrule{1-7}
    $80^\circ$        & $[80^\circ, 89^\circ]$            & 1
                      & 29.60                             & 8.8
                      & 30.23                             & 9.2           \\
                      &                                   & 4
                      & 9.38                              & 10.1
                      & 8.21                              & 10.1          \\
    \bottomrule
  \end{tabular}
  \vspace{-1.0em}
\end{table}

We observed that mixing in a constrained range of larger angles (\eg, $[80^\circ, 85^\circ]$) improves generation quality for both one-step and four-step sampling.
However, including excessively high angles, \eg, $[80^\circ, 89^\circ]$, degrades generation quality and also causes unstable training with gradient explosions.
We therefore conclude that augmenting the training data with a moderate band of large angles enhances quality.

Accordingly, we adopt this mixing strategy for all experiments: for CIFAR-10 with small image size $32$, we add angles from $[80^\circ, 85^\circ]$ with $0.1$ probability to the base range of $[0^\circ, 80^\circ]$;
for ImageNet, Animal-Faces, Oxford-Flowers with large image size $256$, we add angles from $[85^\circ, 89^\circ]$ with $0.1$ probability to the base range of $[0^\circ, 85^\circ]$.

\subsection{Explicit Distribution Regularization}
\label{appsec:explicit_dist_reg}

The ideal distribution on the sphere is uniform.
To form such a distribution, we could add a regularization term to the encoder output.
We investigate two options:
(1) Batch Normalization (BN) \cite{batchnorm} on the encoder output before spherifying;
Since we sample noise from a Normal distribution, BN encourages the encoder output to be Gaussian, which is close to the distribution of noise.
(2) SWD loss \cite{swd,swd-ae,swd-gm,max-swd-gan}, which is a sliced Wasserstein distance between the encoder output and uniform distribution on the sphere.
We implement the loss $\cL_\text{SWD}$ by constructing orthogonal random projections following \cite{swd} Algorithm 3 to the encoder output before spherifying.

Starting with the baseline model without any regularization, we then gradually add BN and SWD loss to evaluate their effects on generation quality.
As shown in \cref{tab:uniform}, applying BN marginally improves generation quality for both one-step and four-step sampling.
However, adding SWD loss on top of BN slightly degrades generation quality.

This suggests that our noisy spherifying method already encourages a near-uniform distribution on the sphere, and additional BN or SWD regularization may not be necessary.
There are downsides to these regularizations as well.
BN introduces extra calibration steps during inference to adjust batch statistics, which complicates the generation process \cite{biggans} (see details in Appendix \ref{appsec:bn_calib}).
And SWD is expensive with large latent dimensions, requiring latent caching and memory for random projection matrices.

\begin{table}[h]
  \centering
  \scriptsize
  \scshape
  \caption{
    \textbf{Ablation on explicit uniform regularization} on the spherical latent space.
  }
  \label{tab:uniform}
  \setlength{\tabcolsep}{4.5pt} 
  \begin{tabular}{@{}lccrcrc@{}} 
    \toprule
                                  &                                   &
                                  & \multicolumn{2}{c}{w/o cfg}
                                  & \multicolumn{2}{c}{w/ cfg $=1.6$}                     \\
    \cmidrule(r){4-5} \cmidrule(l){6-7}
    add-ons                       & steps                             & rFID $\downarrow$
                                  & gFID $\downarrow$                 & IS $\uparrow$
                                  & gFID $\downarrow$                 & IS $\uparrow$     \\
    \midrule
    -                             & 1                                 & 1.20
                                  & 17.56                             & 182.5
                                  & 14.88                             & 217.8             \\
                                  & 4                                 & -
                                  & 8.73                              & 177.6
                                  & 8.05                              & 245.0             \\
    \cmidrule{1-7}
    {\tiny{+}} BN                 & 1                                 & 0.94
                                  & 17.28                             & 181.6
                                  & 14.71                             & 216.0             \\
                                  & 4                                 & -
                                  & 8.39                              & 171.3
                                  & 7.65                              & 240.3             \\
    \cmidrule{1-7}
    {\tiny{+}} $\cL_{\text{SWD}}$ & 1                                 & 1.01
                                  & 18.95                             & 164.4
                                  & 15.47                             & 202.1             \\
                                  & 4                                 & -
                                  & 8.41                              & 165.5
                                  & 7.46                              & 235.2             \\
    \bottomrule
  \end{tabular}
\end{table}

\begin{table}[h]
  \centering
  \scriptsize
  \scshape
  \caption{
    \textbf{Recalibrating BN stats} for inference on ImageNet.
  }
  \label{tab:calib_bn_large}
  \setlength{\tabcolsep}{5pt} 
  \begin{tabular}{@{}lccrcrc@{}}
    \toprule
               &                                   &
               & \multicolumn{2}{c}{w/o cfg}
               & \multicolumn{2}{c}{w/ cfg $=1.6$}                     \\
    \cmidrule(r){4-5} \cmidrule(l){6-7}
    calib by   & steps                             & rFID $\downarrow$
               & gFID $\downarrow$                 & IS $\uparrow$
               & gFID $\downarrow$                 & IS $\uparrow$     \\
    \midrule
    on-the-fly & 1                                 & 0.99
               & 14.21                             & 214.2
               & 13.02                             & 246.0             \\
               & 4                                 & -
               & 6.85                              & 202.3
               & 7.39                              & 268.2             \\
    \cmidrule{1-7}
    generated  & 1                                 & 0.99
               & 14.21                             & 214.2
               & 13.02                             & 246.0             \\
    images     & 4                                 & -
               & 7.30                              & 211.1
               & 8.08                              & 273.9             \\
    \cmidrule{1-7}
    reference  & 1                                 & 1.00
               & 14.21                             & 214.2
               & 13.02                             & 246.0             \\
    images     & 4                                 & -
               & 7.46                              & 209.3
               & 8.06                              & 273.6             \\
    \bottomrule
  \end{tabular}
\end{table}

\subsection{BatchNorm Recalibration}
\label{appsec:bn_calib}

A common issue from BatchNorm \cite{batchnorm} is the train-test discrepancy for the batch statistics, \ie, running mean and variance.
Directly using the training statistics during inference may lead to performance degradation.
Using batch norm, we need to calibrate the stats of bn layers during inference.
DCGANs \cite{dcgans} does not calibarate it.
However, BigGANs \cite{biggans} found the calibration affect the generation quality significantly, and run the testing samples to get the BN stats.

In \cref{tab:calib_bn_large}, we compare the results of using training stats and calibrated stats for BN on ImageNet.
For calibration, we have three options:
(1) no calibration: reset the BN training stats and then directly use the model for generation, the BN stats are updated on-the-fly.
(1) calibrate with reference images: run the model on the training set to accumulate BN statistics prior to generation.
(2) calibrate with generated images: sample $500$ images to get the BN stats before generating the final samples for evaluation.
The results indicate that explicit calibration is not strictly necessary.

\subsection{Volume Compression Ratio}
\label{appsec:com_ratio}

\begin{figure}[t]
  \vspace{1em}
  \centering
  \centerline{\includegraphics[width=1.0\linewidth]{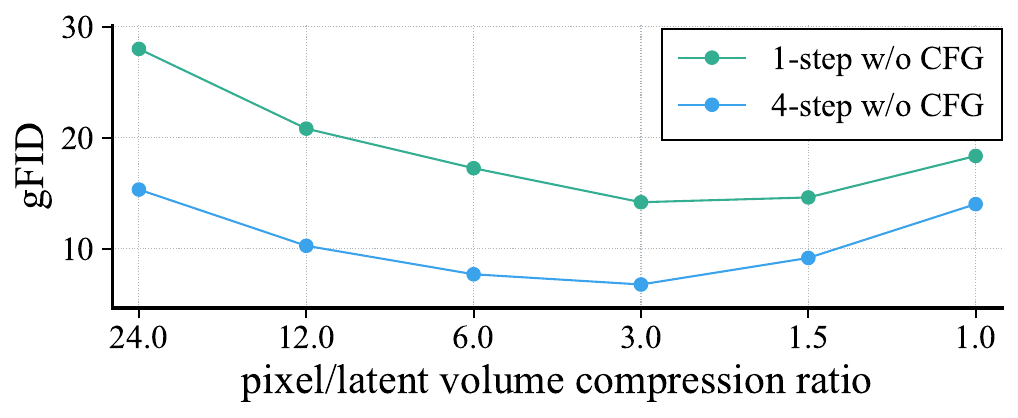}}
  \caption{
    \textbf{Quantitative impact of volume compression ratio} on ImageNet.
    Details in \cref{tab:volume_ratio}.
  }
  \label{fig:volume_ratio}
\end{figure}

Distinct from classical VAEs, our encoder outputs a high latent dimension.
Our pixel/latent volume compression ratio is much smaller than that of a standard VAE, for example, $1.5$ for CIFAR-10 and $3.0$ for ImageNet, compared to $48$ for a standard VAE (\eg, from pixel $256^2\times3$ to latent $32^2\times4$) \cite{ldm,ltx}.
We furthur study the impact of pixel/latent volume compression ratio for the latent dim $L$.
It is well known that, for autoencoders, a higher compression ratio typically leads to worse reconstruction quality but eases diffusion models to fit in \cite{rec-vs-gen,rae}.
However, this is not necessarily true for our method.
We vary the channel depth of latent dim $L$ to adjust the volume compression ratio while keeping the spatial resolution fixed, \ie, $16^2$ for image size $256$ on ImageNet.
The results are plotted in \cref{fig:volume_ratio}, demonstrating two optimal compression ratios around $1.5$ and $3.0$ for few-step generation.
This ratio is way lower than that of typical autoencoders used in diffusion models, \eg, $48$ in \cite{ldm,rae}.
Unless otherwise specified, we use the optimal compression ratio $3.0$ for ImageNet and $1.5$ for CIFAR-10, Animal-Faces and Oxford-Flowers.

\subsection{Noise Prior Distribution}
\label{appsec:noise_prior}

\cref{alg:code} samples noise $\be$ in \cref{eq:generation} from a isotropic Gaussian $\cN(0, I)$ as the ``input latent'' for generation.
Since we spherify it before feeding into the decoder, we assume the generation should be insensitive to the specific choice of noise prior distribution.
This is contrast to GANs \cite{biggan-deep,style-gan-2,stylegan-xl}, which apply truncation tricks \cite{style-gan-1} to the latent prior distribution to improve generation fidelity.
We compare two noise prior distributions: (1) the standard Gaussian $\cN(0, I)$; and (2) a truncated Normal $\cN(0, I)$ truncated to $[-\alpha, \alpha]$, where $\alpha$ is the truncation threshold.
\cref{tab:trunc} confirms the generation quality is insensitive to the choice of noise prior distribution, as the gFID and IS are similar across different $\alpha$ values.

\begin{table}
  \centering
  \scriptsize
  \scshape
  \caption{
    \textbf{Ablation on pixel/latent volume compression ratio} on ImageNet without CFG.
  }
  \label{tab:volume_ratio}
  \setlength{\tabcolsep}{5pt} 
  \begin{tabular}{@{}clccrr@{}} 
    \toprule
    compression ratio & latent $L$       & steps         & rFID $\downarrow$
                      & gFID$\downarrow$ & IS $\uparrow$                     \\
    \midrule
    24                & $16^2\times32$   & 1             & 1.61
                      & 27.99            & 195.6                             \\
                      &                  & 4             & -
                      & 15.34            & 245.1                             \\
    \cmidrule{1-6}
    12                & $16^2\times64$   & 1             & 1.38
                      & 20.82            & 233.8                             \\
                      &                  & 4             & -
                      & 10.28            & 259.0                             \\
    \cmidrule{1-6}
    6.0               & $16^2\times128$  & 1             & 1.15
                      & 17.26            & 225.0                             \\
                      &                  & 4             & -
                      & 7.72             & 229.4                             \\
    \cmidrule{1-6}
    3.0               & $16^2\times256$  & 1             & 1.01
                      & 14.20            & 214.2                             \\
                      &                  & 4             & -
                      & 6.80             & 201.7                             \\
    \cmidrule{1-6}
    1.5               & $16^2\times512$  & 1             & 0.69
                      & 14.63            & 136.0                             \\
                      &                  & 4             & -
                      & 9.19             & 125.9                             \\
    \cmidrule{1-6}
    1.0               & $16^2\times768$  & 1             & 0.64
                      & 18.36            & 97.6                              \\
                      &                  & 4             & -
                      & 14.03            & 85.6                              \\
    \bottomrule
  \end{tabular}
\end{table}
\begin{table}
    \centering
    \scriptsize
    \scshape
    \caption{
        \textbf{Ablation of noise prior distribution} on ImageNet.
    }
    \label{tab:trunc}
    \setlength{\tabcolsep}{4pt} 
    \begin{tabular}{@{}ccrcrc@{}} 
        \toprule
                       &
                       & \multicolumn{2}{c}{w/o cfg}
                       & \multicolumn{2}{c}{w/ cfg $=1.2$}                 \\
        \cmidrule(r){3-4} \cmidrule(l){5-6}
        trunc $\alpha$ & steps
                       & gFID $\downarrow$                 & IS $\uparrow$
                       & gFID $\downarrow$                 & IS $\uparrow$ \\
        \midrule
        -              & 1
                       & 17.81                             & 233.9
                       & 16.30                             & 261.0         \\
                       & 4
                       & 5.92                              & 192.3
                       & 5.02                              & 229.7         \\
        \cmidrule{1-6}
        0.05           & 1
                       & 17.89                             & 231.7
                       & 15.81                             & 277.0         \\
                       & 4
                       & 5.88                              & 192.2
                       & 5.08                              & 228.2         \\
        \cmidrule{1-6}
        0.5            & 1
                       & 17.96                             & 231.4
                       & 15.73                             & 275.8         \\
                       & 4
                       & 5.89                              & 193.5
                       & 5.02                              & 229.7         \\
        \cmidrule{1-6}
        2.0            & 1
                       & 17.82                             & 233.5
                       & 15.66                             & 281.0         \\
                       & 4
                       & 5.85                              & 191.7
                       & 5.06                              & 228.3         \\
        \bottomrule
    \end{tabular}
\end{table}
\begin{table}
    \centering
    \scriptsize
    \scshape
    \caption{
        \textbf{Detailed quantitative results for the impact of the angle $\alpha$} on ImageNet for \cref{fig:sigma_angle_curv}.
    }
    \label{tab:sigma_angle_tbl}
    \setlength{\tabcolsep}{5pt} 
    \begin{tabular}{@{}crccrrrr@{}} 
        \toprule
                       &                                   &                   &
                       & \multicolumn{2}{c}{w/o cfg}
                       & \multicolumn{2}{c}{w/ cfg $=1.6$}                       \\
        \cmidrule(lr){5-6} \cmidrule(l){7-8}
        angle $\alpha$ & $\sigma$
                       & steps                             & rFID $\downarrow$
                       & gFID $\downarrow$                 & IS $\uparrow$
                       & gFID $\downarrow$                 & IS $\uparrow$       \\
        \midrule
        $45^\circ$     & 1.0
                       & 1                                 & 0.69
                       & 69.57                             & 31.3
                       & 59.09                             & 39.9                \\
                       &
                       & 4                                 & -
                       & 80.32                             & 18.7
                       & 73.04                             & 21.9                \\
        \cmidrule{1-8}
        $80^\circ$     & 5.7
                       & 1                                 & 0.87
                       & 60.14                             & 33.4
                       & 51.73                             & 41.3                \\
                       &
                       & 4                                 & -
                       & 42.72                             & 39.3
                       & 34.14                             & 54.3                \\
        \cmidrule{1-8}
        $81^\circ$     & 6.3
                       & 1                                 & 0.94
                       & 47.68                             & 40.4
                       & 40.11                             & 51.1                \\
                       &
                       & 4                                 &
                       & 37.64                             & 43.4
                       & 28.88                             & 62.2                \\
        \cmidrule{1-8}
        $83^\circ$     & 8.1
                       & 1                                 & 0.99
                       & 20.15                             & 107.7
                       & 16.21                             & 135.1               \\
                       &
                       & 4                                 & -
                       & 13.05                             & 102.2
                       & 8.70                              & 151.0               \\
        \cmidrule{1-8}
        $84^\circ$     & 9.5
                       & 1                                 & 1.15
                       & 13.96                             & 161.7
                       & 12.30                             & 186.0               \\
                       &
                       & 4                                 &
                       & 11.92                             & 116.9
                       & 8.66                              & 171.5               \\
        \cmidrule{1-8}
        $85^\circ$     & 11.4
                       & 1                                 & 1.32
                       & 15.40                             & 188.3
                       & 13.14                             & 225.1               \\
                       &
                       & 4                                 & -
                       & 7.53                              & 176.4
                       & 7.28                              & 243.1               \\
        \cmidrule{1-8}
        $86^\circ$     & 14.3
                       & 1                                 & 1.26
                       & 15.37                             & 232.7
                       & 14.15                             & 267.1               \\
                       &
                       & 4                                 & -
                       & 9.96                              & 229.7
                       & 11.14                             & 292.0               \\
        \cmidrule{1-8}
        $87^\circ$     & 19.1
                       & 1                                 & 1.29
                       & 16.40                             & 245.0
                       & 14.94                             & 291.3               \\
                       &
                       & 4                                 & -
                       & 9.99                              & 257.2
                       & 11.50                             & 315.1               \\
        \cmidrule{1-8}
        $88^\circ$     & 28.6
                       & 1                                 & 1.53
                       & 23.27                             & 203.5
                       & 17.72                             & 268.7               \\
                       &
                       & 4                                 & -
                       & 13.15                             & 241.5
                       & 12.24                             & 304.4               \\
        \bottomrule
    \end{tabular}
\end{table}

\section{Hyperparameters}
\label{appsec:hyperparams}

\cref{tab:hyperparams} lists the training hyperparameters and model details for our main experiments.
In addition, for unconditional generation on CIFAR-10, the only difference is removing the class condition and training for $10$K epochs.
We found EMA for smoothing weights in checkpoints may not noticeably change FID or sample quality, likely because we use cosine annealing learning rate schedule, which already provides a smoothing effect.

\begin{table*}[t]
  \centering
  \scriptsize
  \scshape
  \caption{
    \textbf{Training hyperparameters and model details} for main experiments.
  }
  \label{tab:hyperparams}
  \setlength{\tabcolsep}{5pt} 
  \begin{tabular}{@{}lcccc@{}} 
    \toprule
    hyperparameter $\diagdown$ dataset            & cifar-10                  & animal-faces              & oxford-flowers            & imagenet                  \\
    \midrule
    image size                                    & $32$                      & $256$                     & $256$                     & $256$                     \\
    batch size                                    & $256$                     & $256$                     & $256$                     & $256$                     \\
    learning rate (lr)                            & $1 \times 10^{-4}$        & $1 \times 10^{-4}$        & $1 \times 10^{-4}$        & $1 \times 10^{-4}$        \\
    lr decay schedule                             & cosine                    & cosine                    & cosine                    & cosine                    \\
    min lr                                        & $1 \times 10^{-6}$        & $1 \times 10^{-6}$        & $1 \times 10^{-6}$        & $1 \times 10^{-6}$        \\
    weight decay                                  & $0.0$                     & $0.0$                     & $0.0$                     & $0.0$                     \\
    optimizer                                     & adamw                     & adamw                     & adamw                     & adamw                     \\
    warmup epochs                                 & $10$                      & $10$                      & $10$                      & $5$                       \\
    total epochs                                  & $5$K                      & $1$K                      & $1$K                      & $800$                     \\
    \cmidrule{1-5}
    model                                         & sphere-l                  & sphere-l                  & sphere-l                  & sphere-l, -xl             \\
    encoder / decoder size                        & large                     & large                     & large                     & large, xlarge             \\
    number of transformer blocks                  & 24                        & 24                        & 24                        & 24, 28                    \\
    number of attention heads                     & 16                        & 16                        & 16                        & 16                        \\
    transformer hidden size                       & 1024                      & 1024                      & 1024                      & 1024, 1152                \\
    mlp-mixer depth                               & $2$                       & $4$                       & $4$                       & $4$                       \\
    class condition                               & \checkmark                & -                         & \checkmark                & \checkmark                \\
    model params                                  & $921$M                    & $642$M                    & $948$M                    & $950$M                    \\
    volume compression ratio                      & $1.5$                     & $1.5$                     & $1.5$                     & $3.0$                     \\
    latent dim $L$                                & $16^2\times8$             & $32^2\times128$           & $32^2\times128$           & $32^2\times64$            \\
    \cmidrule{1-5}
    angle $\alpha$ jitter range                   & $[\ \ 0^\circ, 80^\circ]$ & $[\ \ 0^\circ, 85^\circ]$ & $[\ \ 0^\circ, 85^\circ]$ & $[\ \ 0^\circ, 85^\circ]$ \\
    angle $\alpha$ mix range                      & $[80^\circ, 85^\circ]$    & $[85^\circ, 89^\circ]$    & $[85^\circ, 89^\circ]$    & $[85^\circ, 89^\circ]$    \\
    angle $\alpha$ mix probability                & $0.1$                     & $0.1$                     & $0.1$                     & $0.1$                     \\
    $\cL_\text{pix-recon}$ smooth l1 loss weight  & $1.0$                     & $25.0$                    & $25.0$                    & $50.0$                    \\
    $\cL_\text{pix-recon}$ perceptual loss weight & $1.0$                     & $1.0$                     & $1.0$                     & $1.0$                     \\
    $\cL_\text{pix-con}$ smooth l1 loss weight    & $0.5$                     & $1.0$                     & $1.0$                     & $25.0$                    \\
    $\cL_\text{pix-con}$ perceptual loss weight   & $0.5$                     & $1.0$                     & $1.0$                     & $1.0$                     \\
    $\cL_\text{lat-con}$ loss weight              & $0.1$                     & $0.1$                     & $0.1$                     & $0.1$                     \\
    \bottomrule
  \end{tabular}
\end{table*}

\clearpage

\section*{Impact Statement}
Our work proposes a new generative framework that offers a fresh perspective on image generation and yields various benefits, including fast sampling and high-dimensional latent space for image generation.
Although this specific method does not raise unique ethical challenges, we acknowledge ongoing general concerns inherent to the field, and we encourage continued communiy discussion to ensure responsible development and mitigation of potential risks.

\bibliography{icml2026}
\bibliographystyle{icml2026}

\end{document}